%% file: Cloud_KSVD_TSP2015.tex
\documentclass[10pt,twocolumn,twoside]{IEEEtran}
\usepackage{cite,graphicx,epstopdf,float,url,amssymb,amsthm,threeparttable,multirow,algorithmic,nameref,paralist}
\usepackage[tight,footnotesize]{subfigure}
\usepackage[ruled,vlined]{algorithm2e}
\usepackage[usenames]{color}
\usepackage[normalem]{ulem}
\usepackage[cmex10]{amsmath}
\IEEEoverridecommandlockouts
\theoremstyle{plain}
\newtheorem{lemma}{Lemma}
\newtheorem{theorem}{Theorem}

\newtheorem{proposition}{Proposition}

\theoremstyle{definition}

\theoremstyle{remark}
\newtheorem{remark}{Remark}

\newcommand{\ConsensusError}{\delta}
\newcommand{\Dictionary}{D}

\newcommand{\tT}{\textsf{T}}

\newcommand{\revtwo}[1]{{\color{black}#1}}

\begin{document}
\title{Cloud K-SVD: A Collaborative Dictionary Learning Algorithm for Big, Distributed Data}
\author{Haroon~Raja and Waheed~U.~Bajwa
\thanks{This work is supported in part by the
ARO under grant W911NF-14-1-0295 and by the NSF under grants CCF-1453073 and
CCF-1525276. Some of the results reported here were presented at Allerton
Conf.~on Communication, Control, and Computing, 2013~\cite{raja2013cloud} and
Intl. Symp. on Information Theory, 2015 ~\cite{rajaconvergence}. The authors
are with the Department of Electrical and Computer Engineering, Rutgers
University--New Brunswick (Emails: {\tt haroon.raja@rutgers.edu} and {\tt
waheed.bajwa@rutgers.edu}).}}

\maketitle

\begin{abstract}
This paper studies the problem of data-adaptive representations for big,
distributed data. It is assumed that a number of
geographically-distributed, interconnected sites have massive local data
and they are interested in collaboratively learning a low-dimensional
geometric structure underlying these data. In contrast to previous works on
subspace-based data representations, this paper focuses on the geometric
structure of a union of subspaces (UoS). In this regard, it proposes a
distributed algorithm---termed cloud K-SVD---for collaborative learning of
a UoS structure underlying distributed data of interest. The goal of cloud
K-SVD is to learn a common overcomplete dictionary at each individual site
such that every sample in the distributed data can be represented through a
small number of atoms of the learned dictionary. Cloud K-SVD accomplishes
this goal without requiring exchange of individual samples between sites.
This makes it suitable for applications where sharing of raw data is
discouraged due to either privacy concerns or large volumes of data. This
paper also provides an analysis of cloud K-SVD that gives insights into its
properties as well as deviations of the dictionaries learned at individual
sites from a centralized solution in terms of different measures of
local/global data and topology of interconnections. Finally, the paper
numerically illustrates the efficacy of cloud K-SVD on real and synthetic
distributed data.
\end{abstract}
\begin{IEEEkeywords}
Consensus averaging, dictionary learning, distributed data, K-SVD, power
method, sparse coding.
\end{IEEEkeywords}

\section{Introduction}\label{sec:Introduction}
Modern information processing is based on the axiom that while real-world
data may live in high-dimensional ambient spaces, relevant information within
them almost always lies near low-dimensional geometric structures. Knowledge
of these (low-dimensional) geometric structures underlying data of interest
is central to the success of a multitude of information processing tasks. But
this knowledge is unavailable to us in an overwhelmingly large number of
applications and a great deal of work has been done in the past to
\emph{learn} geometric structure of data from the data \emph{themselves}.
Much of that work, often studied under rubrics such as \emph{principal
component analysis} (PCA)~\cite{Hotelling.JEP1933}, \emph{generalized PCA}
\cite{Vidal.etal.ITPAMI2005}, \emph{hybrid linear
modeling}~\cite{Zhang.etal.IJCV2012}, and \emph{dictionary learning}
\cite{Kreutz-Delgado.etal.NC2003,aharon2006img,Tosic.Frossard.ISPM2011}, has
been focused on centralized settings in which the entire data are assumed
available at a single location. In recent years, there has been some effort
to extend these works to distributed settings; see, e.g.,
\cite{Gastpar.etal.ITIT2006,Nokleby.Bajwa.Conf2013,li2011distributed,scaglione2008decentralized,macua2010consensus,tron2011distributed,yoon12nips,chen2014dictionary,chainais2013learning,junli2014,chen2014diffusion,chen2014adaptation}.
The setup considered in some of these works is that each distributed entity
is responsible for either some dimensions of the data
\cite{Gastpar.etal.ITIT2006,li2011distributed,Nokleby.Bajwa.Conf2013} or some
part of the learned geometric structure
\cite{li2011distributed,chen2014dictionary,Nokleby.Bajwa.Conf2013}. Other
works in this direction also focus on learning under the assumption of data
lying near \emph{(linear) subspaces}
\cite{Gastpar.etal.ITIT2006,scaglione2008decentralized,macua2010consensus,li2011distributed,Nokleby.Bajwa.Conf2013},
require extensive communications among the distributed entities
\cite{tron2011distributed}, and ignore some of the technical details
associated with processing among distributed entities having interconnections
described by graphs of arbitrary, unknown topologies
\cite{scaglione2008decentralized,macua2010consensus,tron2011distributed,yoon12nips}.

In this paper, we are interested in a setting in which a number of
geographically-distributed sites have massive local data and these sites are
interested in collaboratively learning a geometric structure underlying their
data by communicating among themselves over public/private networks. The key
constraints in this problem that distinguish it from some of the prior works are: ($i$) sites
cannot communicate ``raw'' data among themselves; ($ii$) interconnections
among sites are not described by a complete graph; and ($iii$) sites do not
have knowledge of the global network topology. All these constraints are
reflective of the future of big, distributed data in the world. In
particular, the first constraint is justified because of the size of local
data compilations as well as privacy concerns in the modern age. Similarly,
the latter two constraints are justified because linking
geographically-distributed sites into a complete graph can be cost
prohibitive and since enterprises tend to be protective of their internal
network topologies.

\subsection{Our Contributions}
The first main contribution of this paper is formulation of a distributed
method, which we term as \emph{cloud K-SVD}, that enables data-adaptive
representations in distributed settings. In contrast to works that assume a linear
geometric structure for data
\cite{Gastpar.etal.ITIT2006,scaglione2008decentralized,macua2010consensus,li2011distributed,Nokleby.Bajwa.Conf2013},
cloud K-SVD is based on the premise that data lie near a \emph{union} of
low-dimensional subspaces. The \emph{union-of-subspaces} (UoS) model is a
nonlinear generalization of the subspace model \cite{Lu.Do.ITSP2008} and has
received widespread acceptance in the community lately. The task of learning
the UoS underlying data of interest from data themselves is often termed
\emph{dictionary learning}
\cite{Kreutz-Delgado.etal.NC2003,aharon2006img,Tosic.Frossard.ISPM2011},
which involves data-driven learning of an overcomplete dictionary such that
every data sample can be approximated through a small number of atoms of the
dictionary. Dictionary learning---when compared to linear data-adaptive
representations such as the PCA and the linear discriminant analysis
\cite{Swets.Weng.ITPAMI1996}---has been shown to be highly effective for
tasks such as compression \cite{Kreutz-Delgado.etal.NC2003}, denoising
\cite{elad2006image}, object recognition \cite{Mairal.etal.ITPAMI2012}, and
inpainting \cite{mairal2009online}. Cloud K-SVD, as the name implies, is a
distributed variant of the popular dictionary learning algorithm K-SVD
\cite{aharon2006img} and leverages a classical iterative eigenvector
estimation algorithm, termed the \emph{power
method}~\cite[Ch.~8]{golub2012matrix}, and consensus averaging
\cite{Olfati-Saber.etal.PI2007} for collaborative dictionary learning.

The second main contribution of this paper is a rigorous analysis of cloud
K-SVD that gives insights into its properties as well as deviations of the
dictionaries learned at individual sites from the centralized K-SVD solution
in terms of different measures of local/global data and topology of the
interconnections. Using tools from linear algebra, convex optimization, matrix perturbation
theory, etc., our analysis shows that---under identical initializations---the
dictionaries learned by cloud K-SVD come arbitrarily close to the one learned
by (centralized) K-SVD as long as appropriate number of power method and
consensus iterations are performed in each iteration of cloud K-SVD. Finally,
the third main contribution of this paper involves numerical experiments on
synthetic and real data that demonstrate both the efficacy of cloud K-SVD and
the usefulness of collaborative dictionary learning over local dictionary
learning.
\subsection{Relationship to Previous Work}
\label{subsec:RelatedWork} Some of the earliest works in distributed
processing date back nearly three decades
\cite{Speyer.ITAC1979,tsitsiklis1984convergence}. Since then a number of
distributed methods have been proposed for myriad tasks. Some recent examples
of this that do not involve a centralized \emph{fusion center} include
distributed methods for classification
\cite{forero2010consensus,kokiopoulou2011distributed,Lee.Nedic.IJSTSP2013},
localization \cite{Khan.etal.ITSP2009,Khan.etal.ITSP2010}, linear regression
\cite{mateos2010distributed}, { and (multitask) estimation
\cite{Tu.Sayed.ITSP2012,chen2014diffusion,chen2014adaptation}.} But
relatively little attention has been paid to the problem of data-driven
distributed learning of the geometric structure of data. Notable exceptions
to this include
\cite{scaglione2008decentralized,macua2010consensus,li2011distributed,tron2011distributed,yoon12nips,chen2014dictionary,chainais2013learning,junli2014}.
While our work as well as
\cite{scaglione2008decentralized,macua2010consensus,li2011distributed,tron2011distributed,yoon12nips}
rely on consensus averaging for computing the underlying geometric structure,
we are explicit in our formulation that perfect consensus under arbitrary,
unknown topologies cannot be achieved. In contrast, developments in
\cite{scaglione2008decentralized,macua2010consensus,li2011distributed,tron2011distributed,yoon12nips}
are carried out under the assumption of infinite-precision consensus
averaging. Further,
\cite{scaglione2008decentralized,macua2010consensus,li2011distributed} assume
a subspace data model, while \cite{tron2011distributed} advocates the use of
consensus averaging for computing sample covariance---an approach that
requires extensive communications among the distributed entities.

Our work is most closely related to that
in{~\cite{chen2014dictionary,chainais2013learning,junli2014}},
which also study dictionary learning in distributed settings.
But~\cite{chen2014dictionary} focuses only on learning parts of the
dictionary at each site as opposed to the setup of this paper in which we are
interested in learning a complete dictionary at each site. {While
this paper and~\cite{chainais2013learning,junli2014} share the same setup,
our work as well as \cite{junli2014} are fundamentally different from
\cite{chainais2013learning}. The method proposed
in~\cite{chainais2013learning} involves learning local dictionaries at
different sites and then \emph{diffusing} these local dictionaries to obtain
a global dictionary. In contrast, our work and~\cite{junli2014} are based on
the centralized K-SVD algorithm, which is known to be superior to other
dictionary learning methods \cite{aharon2006img}, and involve updating each
atom of the local dictionaries in a collaborative fashion.
The difference between this work and~\cite{junli2014} lies in the fact that
cloud K-SVD uses a distributed variant of the power method to update each
atom, whereas \cite{junli2014} relies on distributed optimization for this
purpose. This helps us rigorously analyze the performance of cloud K-SVD,
whereas no such analysis is provided in~\cite{junli2014}. Note that while we
analyzed the distributed power method component of cloud K-SVD in our earlier
work~\cite{raja2013cloud}, this paper extends that work to provide a
comprehensive analysis of the entire algorithm.}

We conclude by noting that the distributed power method component of cloud K-SVD has similarities with the work
in~\cite{scaglione2008decentralized,yildiz2009distributed}. However, unlike~\cite{scaglione2008decentralized,yildiz2009distributed}, we do not
assume perfect consensus during iterations of the power method, which leaves
open the question of convergence of the distributed variant of the power
method. While analyzing cloud K-SVD, we in fact end up addressing this
question also. That part of our analysis is reminiscent of the one carried
out in \cite{kempe2008decentralized} in the context of convergence behavior
of distributed eigenanalysis of a network using a power method-like iterative
algorithm. However, there are fundamental differences in the analysis of
\cite{kempe2008decentralized} and our work because of the exact place where
consensus averaging is carried out in the two works, which is dictated by the
distinct nature of the two applications.
\subsection{Notation and Paper Organization}
We use lower-case letters to represent scalars and vectors, while we use
upper-case letters to represent matrices. The operator $\textrm{sgn} :
\mathbb{R} \rightarrow \{+1,-1\}$ is defined as $\textrm{sgn}(x) = x/|x|$,
while $\textrm{supp}(v)$ returns indices of the nonzero entries in vector
$v$. Superscript $(\cdot)^\tT$ denotes the transpose operation, $\|\cdot\|_0$
counts the number of nonzero entries in a vector, $\|v\|_p$ denotes the usual
$\ell_p$ norm of vector $v$, and $\langle u, v \rangle$ denotes the inner
product between vectors $u$ and $v$. Given a set $\mathcal{I}$,
$v_{|\mathcal{I}}$  and $A_{|\mathcal{I}}$ denote a subvector and a submatrix
obtained by retaining entries of vector $v$ and columns of matrix $A$
corresponding to the indices in $\mathcal{I}$, respectively, while $\|A\|_2$,
$\|A\|_F$, and $\|A\|_{\max}$ denote the operator norm, Frobenius norm, and
max norm (i.e., maximum absolute value) of matrix $A$, respectively.
{Given matrices $\{A_i\in\mathbb{R}^{n_i\times m_i}\}_{i=1}^{N}$,
$\text{diag}\{A_1,\dots, A_N\}$ denotes a block-diagonal matrix $A\in
\mathbb{R}^{\sum{n_i}\times \sum{m_i}}$ that has $A_i$'s on its diagonal.}
Finally, given a matrix $A$, $a_j$ and $a_{j,T}$ denote the $j^{th}$ column
and the $j^{th}$ row of $A$, respectively.

The rest of this paper is organized as follows. In
Sec.~\ref{section:ProblemFormulation}, we formulate the problem of
collaborative dictionary learning from big, distributed data. In
Sec.~\ref{section:ProposedAlgorithm}, we describe the cloud K-SVD
algorithm. In Sec.~\ref{section:Analysis}, we provide an analysis of cloud
K-SVD algorithm. We provide some numerical results in
Sec.~\ref{section:NumericalEvaluation} and concluding remarks in
Sec.~\ref{section:conclusion}. Finally, proofs of main theorems stated in Sec.~\ref{section:Analysis} are given in appendices.
%

\input{docs/ProblemFormulation}
\input{docs/MethodologyV2}
\input{docs/ConvergenceAnalysisV2}
\input{docs/NumericalEvaluationV2}
\input{docs/Conclusion}
\input{docs/Appendix}

\end{document}

%% file: docs/ProblemFormulation.tex
\section{Problem Formulation}\label{section:ProblemFormulation}
In this paper, we consider a collection of $N$ geographically-distributed
sites that are interconnected to each other according to a fixed topology.
Here, we use ``site'' in the broadest possible sense of the term, with a site
corresponding to a single computational system (e.g., sensor, drone,
smartphone, tablet, server, database), a collection of co-located
computational systems (e.g., data center, computer cluster, robot swarm),
etc. Mathematically, we represent this collection and their interconnections
through an undirected graph $\mathcal{G}=(\mathcal{N},\mathcal{E})$, where
$\mathcal{N}=\{1,2,\cdots,N\}$ denotes the sites and $\mathcal{E}$ denotes
edges in $\mathcal{G}$ with $(i,i) \in \mathcal{E}$, while $(i,j) \in
\mathcal{E}$ whenever there is a connection between sites $i$ and $j$. The
only assumption we make about the topology of $\mathcal{G}$ is that it is a
connected graph.

\begin{figure}[t]
\centering
\includegraphics[width=\columnwidth]{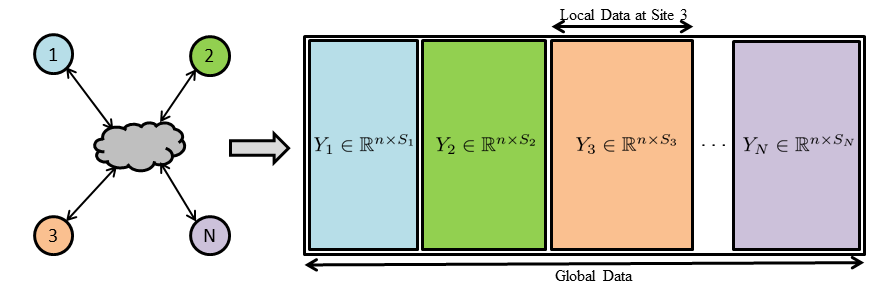}%
\caption{A schematic representing global data $Y$ distributed across $N$ sites. Here, $n$ denotes the dimension of each data sample, while $S_i$ denotes the total number of data samples available at the $i^{th}$ site.}%
\label{fig:DataRepresentation}%
\vspace{-\baselineskip}%
\end{figure}

Next, we assume each site $i$ has a collection of local data,
expressed as a matrix $Y_i \in \mathbb{R}^{n \times S_i}$ with
 $S_i$ representing the number of data samples at the $i^{th}$ site. We can express
all this distributed data into a single matrix {\small$Y =
\begin{bmatrix}Y_1 & \dots & Y_N\end{bmatrix} \in \mathbb{R}^{n \times S}$,}
where $S = \sum_{i=1}^{N} S_i$ denotes the total number of data samples
distributed across the $N$ sites; see Fig.~\ref{fig:DataRepresentation} for a
schematic representation of this. In this setting, the fundamental objective
is for each site to collaboratively learn a low-dimensional geometric
structure that underlies the global (distributed) data $Y$. The basic
premises behind collaborative structure learning of global data, as opposed
to local structure learning of local data, are manifold. First, since the
number of global samples is much larger than the number of local samples, we
expect that collaborative learning will outperform local learning for data
representations. Second, local learning will be strictly suboptimal for some
sites in cases where sampling density, noise level, fraction of outliers,
etc., are not uniform across all sites. Collaborative learning, on the other
hand, will even out such nonuniformities within local data.

Our main assumption is that the low-dimensional geometric structure
underlying the global data corresponds to a union of $T_0$-dimensional
subspaces in $\mathbb{R}^n$, where $T_0 \ll n$. One possible means of
learning such a structure is studied under the moniker dictionary learning,
which learns an \emph{overcomplete dictionary} $D$ such that each data sample
is well approximated by no more than $T_0$ columns (i.e., \emph{atoms}) of
$D$ \cite{Kreutz-Delgado.etal.NC2003,aharon2006img,Tosic.Frossard.ISPM2011}.
Assuming the global data $Y$ is available at a centralized location, this
problem of dictionary learning can be expressed as
\begin{align}
\label{eq:MainEquation}
\big(D, X\big)=\arg\min_{D,X}\|Y-\Dictionary X\|_{F}^{2} \ \text{s.t.} \ \forall s, \|x_{s}\|_{0} \leq T_{0},
\end{align}
where $D \in \mathbb{R}^{n \times K}$ with $K > n$ is an overcomplete
dictionary having unit $\ell_2$-norm columns, $X \in \mathbb{R}^{K \times S}$
corresponds to representation coefficients of the data having no more than
$T_0 \ll n$ nonzero coefficients per sample, and $x_s$ denotes the $s^{th}$
column in $X$. Note that \eqref{eq:MainEquation} is non-convex in $\big(D,
X\big)$, although it is convex in $D$ alone. One of the most popular
approaches to solving \eqref{eq:MainEquation} involves alternate minimization
in which one alternates between solving \eqref{eq:MainEquation} for $D$ using
a fixed $X$ and then solving \eqref{eq:MainEquation} for $X$ using a fixed
$D$ \cite{engan2000multi,aharon2006img}.

Unlike classical dictionary learning, however, we do not have the global data
$Y$ available at a centralized location. Data aggregation either at a
centralized location or at any one of the individual sites is also
impractical due to communications and storage costs of big data. Furthermore, privacy issues may also preclude aggregation of data. Instead, our
goal is to have individual sites collaboratively learn dictionaries
$\{\widehat{D}_i\}_{i\in\mathcal{N}}$ from global data $Y$ such that these
\emph{collaborative dictionaries} are close to a dictionary $D$ that could
have been learned from $Y$ in a centralized fashion. In the following section,
we present a distributed variant of a popular dictionary learning algorithm
that accomplishes this goal without exchanging raw data between sites. This is followed by a rigorous analysis of the
proposed algorithm in Sec.~\ref{section:Analysis}, which establishes that the
collaborative dictionaries learned using our proposed algorithm can indeed be
made to come arbitrarily close to a centralized dictionary.

%% file: docs/MethodologyV2.tex
\section{Cloud K-SVD}
\label{section:ProposedAlgorithm} In this paper, we focus on the K-SVD
algorithm \cite{aharon2006img} as the basis for collaborative dictionary
learning. We have chosen to work with K-SVD because of its  iterative nature
and its reliance on the singular value decomposition (SVD), both of which
enable its exploitation for distributed purposes. In the following, we first
provide a brief overview of K-SVD, which is followed by presentation of our
proposed algorithm---termed cloud K-SVD---for collaborative dictionary
learning.
\subsection{Dictionary Learning Using K-SVD}
\label{subsec:reviewKSVD} The K-SVD algorithm initializes with a (often
randomized) dictionary $D^{(0)}$ and solves \eqref{eq:MainEquation} by iterating
between two stages: a \emph{sparse coding stage} and a \emph{dictionary
update stage} \cite{aharon2006img}. Specifically, for a fixed estimate of the
dictionary $D^{(t-1)}$ at the start of iteration $t\geq 1$, the sparse coding stage in K-SVD involves solving
\eqref{eq:MainEquation} for $X^{(t)}$ as follows:
\begin{align}
\label{eq:sparseCoding}
 \forall s, \ x_s^{(t)} = \arg\min_{x \in \mathbb{R}^K}{\|y_s- D^{(t-1)} x\|_{2}^{2}}\; \text{s.t.} \; \|x\|_0 \leq T_0,
\end{align}
where $y_s$ denotes the $s^{th}$ column of $Y$. While \eqref{eq:sparseCoding}
in its stated form has combinatorial complexity, it can be solved by either
convexifying \eqref{eq:sparseCoding} \cite{chen1998atomic} or using greedy
algorithms \cite{tropp2007signal}.

After the sparse coding stage, K-SVD fixes $X^{(t)}$ and moves to the
dictionary update stage. The main novelty in K-SVD lies in the manner in
which it carries out dictionary update, which involves iterating through the
$K$ atoms of $D^{(t-1)}$ and individually updating the $k^{th}$ atom,
$k\in{1,\dots,K}$, as follows:
\begin{align} \label{eq:Centralized_K_SVD}
d_k^{(t)} &= \arg\min_{d \in \mathbb{R}^n} \Bigg\|\Big(Y-\sum_{j=1}^{k-1}{d_{j}^{(t)}x_{j,T}^{(t)}}-\sum_{j=k+1}^{K}{d_{j}^{(t-1)} x_{j,T}^{(t)}}\Big)\nonumber\\
&\qquad\quad-d\,x_{k,T}^{(t)}\Bigg\|_{F}^{2} =\arg\min_{d \in \mathbb{R}^n} \left\|E_{k}^{(t)}-d\,x_{k,T}^{(t)}\right\|_{F}^{2}.
\end{align}
Here, $E_k^{(t)}$ is the representation error for $Y$ using first $k-1$ atoms
of $D^{(t)}$ and last $k+1,\dots,K$ atoms of $D^{(t-1)}$. In order to
simplify computations, K-SVD in \cite{aharon2006img} further defines an
ordered set $\omega_k^{(t)} = \{s: 1 \leq s \leq S, x_{k,T}^{(t)}(s) \neq
0\}$, where $x_{k,T}^{(t)}(s)$ denotes the $s^{th}$ element of
$x_{k,T}^{(t)}$, and an $S \times |\omega_k^{(t)}|$ binary matrix
$\Omega_k^{(t)}$ that has ones in $(\omega_k^{(t)}(s), s)$ locations and
zeros everywhere else. Then, defining $E_{k,R}^{(t)}=E_k^{(t)}
\Omega_k^{(t)}$ and $x_{k,R}^{(t)} = x_{k,T}^{(t)}\Omega_k^{(t)}$, it is easy
to see from \eqref{eq:Centralized_K_SVD} that $d_k^{(t)} = \arg\min_{d \in
\mathbb{R}^n}\left\|E_{k,R}^{(t)}-d\,x_{k,R}^{(t)}\right\|_{F}^{2}$.
Therefore, solving \eqref{eq:Centralized_K_SVD} is equivalent to finding the
best rank-one approximation of $E_{k,R}^{(t)}$, which is given by the
Eckart--Young theorem as $d_{k}^{(t)} x_{k,R}^{(t)}= \sigma_1 u_1
v_{1}^{\tT},$ where $u_1$ and $v_1$ denote the largest left- and
right-singular vectors of $E_{k,R}^{(t)}$, respectively, while $\sigma_1$
denotes the largest singular value of $E_{k,R}^{(t)}$. The $k^{th}$ atom of
$D^{(t)}$ can now simply be updated as $d_k^{(t)} = u_1$. It is further
advocated in \cite{aharon2006img} that the $k^{th}$ row of the ``reduced''
coefficient matrix, $x_{k,R}^{(t)}$, should be simultaneously updated to
$x_{k,R}^{(t)} = \sigma_1 v_1^\tT$. The dictionary update stage in K-SVD
involves $K$ such applications of the Eckart--Young theorem to update the $K$
atoms of $D^{(t-1)}$ and the $K$ ``reduced'' rows of $X^{(t)}$. The algorithm
then moves to the sparse coding stage and continues alternating between the
two stages till a stopping criterion (e.g., a prescribed representation
error) is reached.

\subsection{Collaborative Dictionary Learning Using Cloud K-SVD}
\label{subsec:Distributed_K-SVD} We now present our collaborative dictionary
learning algorithm based on K-SVD. The key to distributing K-SVD is
understanding ways in which both the sparse coding and the dictionary update
stages can be distributed. To this end, we assume collaborative dictionary
learning is in iteration $t \geq 1$ and each site $i$ in this iteration has a
local estimate $\widehat{D}_i^{(t-1)}$ of the desired dictionary from the
previous iteration. In order for the sparse coding stage to proceed, we
propose that each site computes representation coefficients of its local data
without collaborating with other sites by locally solving Step 3 of Algorithm~\ref{algo:Distributed_KSVD}, i.e.,
\begin{equation}
\forall s, \ \widetilde{x}_{i,s}^{(t)} = \arg\min_{x \in \mathbb{R}^K}{\|y_{i,s}-\widehat{D}_{i}^{(t-1)} x\|_{2}^{2} \; \text{s.t.} \; \|x\|_0 \leq T_0},
\label{eq:DistributedSparseCoding}
\end{equation}
where $y_{i,s}$ and $\widetilde{x}_{i,s}^{(t)}$ denote the $s^{th}$ sample
and its coefficient vector at site $i$, respectively. This ``local'' sparse
coding for collaborative dictionary learning simplifies the sparse coding
stage and is justified as long as the local dictionary estimates
$\widehat{D}_i^{(t-1)}$ remain close to each other (established in
Sec.~\ref{section:Analysis}).


The next challenge in collaborative dictionary learning based on K-SVD arises
during the dictionary update stage. Recall that the dictionary update stage
in K-SVD involves computing the largest left- and right-singular vectors of
the ``reduced'' error matrix $E_{k,R}^{(t)} = E_k^{(t)} \Omega_k^{(t)}, k\in
\{1,\dots,K\}$. However, unless the local dictionary estimates
$\widehat{D}_i^{(t-1)}$ happen to be identical, we end up with $N$ such
(reduced) error matrices in a distributed setting due to $N$ different local
dictionary estimates. To resolve this, we propose to use the following
definition of the reduced error matrix in a distributed setting:
$\widehat{E}_{k,R}^{(t)} =
\begin{bmatrix} \widehat{E}_{1,k,R}^{(t)} & \dots &
\widehat{E}_{N,k,R}^{(t)}\end{bmatrix},$ where
{\small$\widehat{E}_{i,k,R}^{(t)}= Y_{i}\widetilde{\Omega}_{i,k}^{(t)} -
\Big(\sum_{j=1}^{k-1}{\widehat{d}_{i,j}^{(t)}
\widehat{x}_{i,j,T}^{(t)}}+\sum_{j=k+1}^{K}{\widehat{d}_{i,j}^{(t-1)}
\widetilde{x}_{i,j,T}^{(t)}}\Big)\widetilde{\Omega}_{i,k}^{(t)}$}. Here,
$\widetilde{x}_{i,j,T}^{(t)}$ denotes the $j^{th}$ row of coefficient matrix
$\widetilde{X}_i^{(t)}$ computed at site $i$ during the sparse coding step
performed on $Y_i$ using $\widehat{D}_i^{(t-1)}$ at the start of iteration
$t$, while $\widehat{x}_{i,j,T}^{(t)}$ denotes the $j^{th}$ row of the
updated coefficient matrix $\widehat{X}_i^{(t)}$ available at site $i$ due to
the update in coefficient matrix performed during the dictionary update step.
Furthermore, $\widetilde{\Omega}_{i,k}^{(t)}$ is similar to $\Omega_k^{(t)}$
defined for K-SVD except that it is now defined for only local coefficient
matrix $\widetilde{X}_i^{(t)}$ at site $i$.


Next, in keeping with the K-SVD derivation in \cite{aharon2006img}, we
propose that each of the $N$ sites updates the $k^{th}$ atom of its
respective local dictionary and the $k^{th}$ row of its respective
``reduced'' coefficient matrix, $\widehat{x}_{i,k,R}^{(t)} =
\widehat{x}_{i,k,T}^{(t)}\widetilde{\Omega}_{i,k}^{(t)}$, by collaboratively
computing the dominant left- and right-singular vectors of the
\emph{distributed error matrix} $\widehat{E}_{k,R}^{(t)}$, denoted by $u_1$
and $v_1$, respectively.{\footnote{An alternative is to compute an estimate of $\widehat{E}_{k,R}^{(t)}$
at each site using consensus averaging, after which individual sites can compute SVD of
$\widehat{E}_{k,R}^{(t)}$ locally. Despite its apparent simplicity, this
approach will have significantly greater communication overhead compared to
our proposed method.}} In fact, since $u_1^\tT \widehat{E}_{k,R}^{(t)} =
\sigma_1 v_1$ with $\sigma_1$ being the largest singular value of
$\widehat{E}_{k,R}^{(t)}$, it follows that if a site has access to the
dominant left-singular vector, $u_1$, of $\widehat{E}_{k,R}^{(t)}$ then it
can simply update the $k^{th}$ row of its respective ``reduced'' coefficient
matrix by setting $\widehat{d}_{i,k}^{(t)} = u_1$ and setting
$\widehat{x}_{i,k,R}^{(t)} = \widehat{d}_{i,k}^{(t)^{\tT}}
\widehat{E}_{i,k,R}^{(t)}$. Therefore, we need only worry about collaborative
computation of $u_1$ in this setting. To this end, we define
$\widehat{M}^{(t)} = \widehat{E}_{k,R}^{(t)} \widehat{E}_{k,R}^{(t)^{\tT}}$
and note that $u_1$ corresponds to the dominant eigenvector of
$\widehat{M}^{(t)}$. Now express $\widehat{M}^{(t)}$ as $\widehat{M}^{(t)}
=\sum_{i=1}^{N} \widehat{M}_i^{(t)}$ and notice that each
$\widehat{M}_i^{(t)} = \widehat{E}_{i,k,R}^{(t)}
\widehat{E}_{i,k,R}^{(t)^{\tT}}$ is a matrix that is readily computable at
each local site. Our goal now is computing the dominant eigenvector of
$\widehat{M}^{(t)} = \sum_{i=1}^{N} \widehat{M}_i^{(t)}$ in a collaborative
manner at each site. In order for this, we will make use of distributed power
method, which has been invoked previously in \cite{kempe2008decentralized,
scaglione2008decentralized, macua2010consensus} and which corresponds to a
distributed variant of the classical power method for eigenanalysis
\cite{golub2012matrix}.

\textbf{\emph{Distributed Power Method:}} Power method is an iterative
procedure for computing eigenvectors of a matrix. It is simple to implement
and, assuming that the largest eigenvalue $\lambda_1$ of a matrix is strictly
greater than its second-largest eigenvalue $\lambda_2$, it converges to the
subspace spanned by the dominant eigenvector at an exponential rate. We are
interested in a distributed variant of the power method to compute the
dominant eigenvector of $\widehat{M}^{(t)} = \sum_{i=1}^N
\widehat{M}_i^{(t)}$, where the $\widehat{M}_i^{(t)}$'s are distributed
across $N$ sites. To this end, we proceed as follows.

First, all sites initialize to the same (unit-norm) estimate of the
eigenvector $\widehat{q}_i^{(0)} = q^{init}$.\footnote{This can be
accomplished, for example, through the use of (local) random number
generators initialized with the same seed. Also, note that a key requirement
in power method is that $\langle u_1, q^{init}\rangle \not= 0$, which is
ensured with very high probability in the case of a random initialization.}
Next, assuming that the sites are carrying out iteration $t_p$ of the
distributed power method, each site computes $\widehat{M}_i^{(t)}
\widehat{q}_i^{(t_p-1)}$ locally, where $\widehat{q}_i^{(t_p-1)}$ denotes an
estimate of the dominant eigenvector of $\widehat{M}^{(t)}$ at the $i^{th}$
site after $t_p-1$ power method iterations. In the next step, the sites collaboratively
compute an approximation $\widehat{v}_{i}^{(t_p)}$ of $\sum_i
\widehat{M}_i^{(t)} \widehat{q}_i^{(t_p-1)}$ at each site. In the final step
of the $t_p^{th}$ iteration of the distributed power method, each site
normalizes its estimate of the dominant eigenvector of $\widehat{M}^{(t)}$
locally: $\widehat{q}_i^{(t_p)}
=\widehat{v}_{i}^{(t_p)}/\|\widehat{v}_{i}^{(t_p)}\|_2$.

It is clear from the preceding discussion that the key in distributed power
method is the ability of the sites to collaboratively compute an
approximation of $\sum_i \widehat{M}_i^{(t)} \widehat{q}_i^{(t_p-1)}$ in each
iteration. In order for this, we make use of the popular consensus averaging
method \cite{xiao2004fast}. To perform consensus averaging, we first design a
doubly-stochastic weight matrix $W$ that adheres to the topology of the
underlying graph $\mathcal{G}$. In particular, we have that $w_{i,j} = 0$
whenever $(i,j) \not\in \mathcal{E}$. We refer the reader to
\cite{olshevsky2009convergence, xiao2004fast, olfati2004consensus} for
designing appropriate weight matrices in a distributed manner without relying
on knowledge of the global network topology. In order to compute $\sum_i
\widehat{M}_i^{(t)} \widehat{q}_i^{(t_p-1)}$ using consensus averaging, each
site is initialized with $z_i^{(0)} = \widehat{M}_i^{(t)}
\widehat{q}_{i}^{(t_p-1)}$. Next, let $\mathcal{N}_i = \{j : (i,j) \in
\mathcal{E}\}$ be the neighborhood of site $i$, define ${Z^{(0)}}
=\begin{bmatrix} z_1^{(0)} & \dots &z_N^{(0)}\end{bmatrix}^{\tT}$, and assume
we are in $t_c^{th}$ iteration of consensus averaging. Then consensus works
by having each site carry out the following updates in each consensus
iteration through communications with its neighbors: $z_i^{(t_c)}=\sum_{j \in
\mathcal{N}_i}{w_{i,j}z_{j}^{(t_c-1)}}$. The dynamics of the overall system
in this case evolve as $Z^{(t_c)}=W^{t_c}Z^{(0)}.$ It then follows that
$Z^{(t_c)}_{i,T} \stackrel{t_c}{\longrightarrow} \mathbf{1}^{\tT}Z^{(0)}/N$
\cite{xiao2004fast}, where $Z^{(t_c)}_{i,T}$ denotes the $i^{th}$ row of
$Z^{(t_c)}$ and $\mathbf{1}\in\mathbb{R}^N$ denotes a (column) vector of all
ones. This in particular implies that each site achieves perfect consensus
averaging as $t_c \rightarrow \infty$ and obtains $
{Z^{(\infty)}_{i,T}}^{\tT} = \frac{1}{N} \sum_{j=1}^{N}z_{j}^{(0)} =
\frac{1}{N} \sum_{j=1}^{N} \widehat{M}_j^{(t)} \widehat{q}_j^{(t_p-1)}$.

But one can not perform infinite consensus iterations in practice within each
iteration of the distributed power method. Instead, we assume a finite number
of consensus iterations, denoted by $T_c$, in each power method iteration and
make use of the modification of standard consensus averaging proposed in
\cite{kempe2008decentralized} to obtain $\widehat{v}_{i}^{(t_p)} =
{Z^{(T_c)}_{i,T}}^{\tT}/[W^{T_c}_1]_i$, where $W^{T_c}_1$ is the first column
of $W^{T_c}$ and $[\cdot]_i$ denotes the $i^{th}$ entry of a vector. Note
that this leads to an error $\epsilon_{i,c}^{(t_p)}$ within
$\widehat{v}_{i}^{(t_p)}$ at each site for any finite $T_c$, i.e.,
$\widehat{v}_i^{(t_p)} = {Z^{(T_c)}_{i,T}}^{\tT}/[W^{T_c}_1]_i =
\sum_{j=1}^{N} \widehat{M}_j\widehat{q}_j^{(t_p-1)} +
\epsilon_{i,c}^{(t_p)}.$ After finishing consensus iterations, each site $i$
in iteration $t_p$ of power method normalizes this vector
$\widehat{v}_{i}^{(t_p)}$ to get an estimate of the dominant eigenvector of
$\widehat{M}^{(t)}$. Finally, we carry out enough iterations of the
distributed power method at each site that the error between successive
estimates of the eigenvector falls below a prescribed threshold.

\begin{algorithm}[t]
\textbf{Input:} Local data $Y_{1}, Y_{2},\ldots,Y_{N}$, problem parameters $K$ and $T_0$, and doubly-stochastic matrix $W$.\\
\textbf{Initialize:} Generate $d^{ref} \in \mathbb{R}^n$ and
$\Dictionary^{init} \in \mathbb{R}^{n \times K}$ randomly, set $t\leftarrow
0$ and $\widehat{\Dictionary}_{i}^{(t)} \leftarrow \Dictionary^{init},
i=1,\ldots,N$.%
\vspace{-\baselineskip}%
\algsetup{indent=1em}
\begin{algorithmic}[1]
\WHILE{$stopping~\, rule$}%
\STATE $t\leftarrow t+1$%
\STATE (\textit{Sparse Coding}) The $i^{th}$ site solves $\forall s,
{\widetilde{x}_{i,s}^{(t)} \leftarrow \arg\min\limits_{x \in
\mathbb{R}^{K}}{\|y_{i,s}-\widehat{\Dictionary}_{i}^{(t-1)} x\|_{2}^{2}
\ \text{s.t.} \ \|x\|_0 \leq T_0}}$%
\FOR{$k=1 \; \TO \; K$ (\emph{Dictionary Update})}%
\STATE $\widehat{E}_{i,k,R}^{(t)}\leftarrow
Y_{i}\widetilde{\Omega}_{i,k}^{(t)}~-\sum_{j=1}^{k-1}{\widehat{d}_{i,j}^{(t)}
\widehat{x}_{i,j,T}^{(t)}\widetilde{\Omega}_{i,k}^{(t)}}$\\\qquad\qquad\qquad\qquad\qquad$-\sum_{j=k+1}^{K}{\widehat{d}_{i,j}^{(t-1)}\widetilde{x}_{i,j,T}^{(t)}\widetilde{\Omega}_{i,k}^{(t)}}$%
\STATE $\widehat{M}_i\leftarrow
\widehat{E}^{(t)}_{i,k,R}{\widehat{E}_{i,k,R}}^{(t)^{\tT}}$
\STATE \textbf{(\emph{Initialize Distributed Power Method})} Generate
$q^{init}$ randomly, set $t_p \leftarrow 0$ and $\widehat{q}_{i}^{(t_p)}
\leftarrow
q^{init}$%
\WHILE{$stopping~\, rule$}%
\STATE $t_p\leftarrow t_p+1$%
\STATE \textbf{(\emph{Initialize Consensus Averaging})} Set $t_c \leftarrow
0$ and $z_{i}^{(t_c)}\leftarrow \widehat{M}_i \widehat{q}_{i}^{(t_p-1)}$%
\WHILE{$stopping~\, rule$}%
\STATE $t_c \leftarrow t_c+1$%
\STATE $z_i^{(t_c)}\leftarrow\sum_{j \in \mathcal{N}_i}{w_{i,j}z_{i}^{(t_c-1)}}$%
\ENDWHILE%
\STATE $\widehat{v}_{i}^{(t_p)} \leftarrow z_{i}^{(t_c)}/[W^{t_c}_1]_i$%
\STATE $\widehat{q}_{i}^{(t_p)} \leftarrow
\widehat{v}_{i}^{(t_p)}/\|\widehat{v}_{i}^{(t_p)}\|_2$%
\ENDWHILE%
\STATE $\widehat{d}_{i,k}^{(t)} \leftarrow \textrm{sgn}\left(\langle d^{ref},
\widehat{q}_{i}^{(t_p)}\rangle\right) \widehat{q}_{i}^{(t_p)}$%
\STATE $\widehat{x}_{i,k,R}^{(t)} \leftarrow
\widehat{d}_{i,k}^{(t)^{\tT}}\widehat{E}_{i,k,R}^{(t)}$%
\ENDFOR%
\ENDWHILE%
\end{algorithmic}
{\bf Return:} $\widehat{\Dictionary}_{i}^{(t)}, i=1,2,\dots,N$.%
\caption{ Cloud K-SVD for dictionary learning} \label{algo:Distributed_KSVD}
\end{algorithm}
We have now motivated and described the key components of our proposed
algorithm and the full collaborative dictionary learning algorithm, termed
\emph{cloud K-SVD}, is detailed in Algorithm~\ref{algo:Distributed_KSVD}.
Notice the initialization of cloud K-SVD differs from K-SVD in the sense that
each site also generates a common (random) reference vector $d^{ref} \in
\mathbb{R}^n$ and stores it locally. The purpose of $d^{ref}$ is to ensure
that the eigenvectors computed by different sites using the distributed power
method all point in the same quadrant, rather than in antipodal quadrants
(Step~18 in Algorithm~\ref{algo:Distributed_KSVD}). While this plays a role
in analysis, it does not have an effect on the workings of cloud K-SVD.
Notice also that we have not defined any stopping rules in
Algorithm~\ref{algo:Distributed_KSVD}. One set of rules could be to run the
algorithm for fixed dictionary learning iterations $T_d$, power method
iterations $T_p$, and consensus iterations $T_c$. It is worth
noting here that algorithms such as cloud K-SVD are often referred to as two
time-scale algorithms in the literature. Nonetheless, cloud K-SVD with the
stopping rules of finite $(T_d, T_p, T_c)$ can be considered a \emph{quasi}
one time-scale algorithm. Accordingly, our analysis of cloud K-SVD assumes
these stopping rules.

\begin{remark}
A careful reading of Algorithm~\ref{algo:Distributed_KSVD} reveals that
normalization by $[W^{t_c}_1]_i$ in Step~15 is redundant due to the
normalization in Step~16. We retain the current form of Step~15 however to
facilitate the forthcoming analysis.
\end{remark}

%
%

%% file: docs/ConvergenceAnalysisV2.tex
\section{Analysis of Cloud K-SVD}
\label{section:Analysis} Since power method and consensus averaging in
Algorithm~\ref{algo:Distributed_KSVD} cannot be performed for an infinite
number of iterations, in practice this results in residual errors in each iteration of the algorithm. It is therefore important to understand
whether the dictionaries $\{\widehat{D}_i\}$ returned by cloud K-SVD approach
the dictionary that could have been obtained by centralized K-SVD
\cite{aharon2006img}. In order to address this question, we need to
understand the behavior of major components of cloud K-SVD, which include
sparse coding, dictionary update, and distributed power method within
dictionary update. In addition, one also expects that the closeness of
$\widehat{D}_i$'s to the centralized solution will be a function of certain
properties of local/global data. We begin our analysis of cloud K-SVD by
first stating some of these properties in terms of the centralized K-SVD
solution.

\subsection{Preliminaries}
\label{ssec:ksvd_prelim} The first thing needed to quantify deviations of the
cloud K-SVD dictionaries from the centralized K-SVD dictionary is algorithmic
specification of the sparse coding steps in both algorithms. While the sparse
coding steps as stated in \eqref{eq:sparseCoding} and
\eqref{eq:DistributedSparseCoding} have combinatorial complexity, various
low-complexity computational approaches can be used to solve these steps in
practice. Our analysis in the following will be focused on the case when
sparse coding in both cloud K-SVD and centralized K-SVD is carried out using
the \emph{lasso} \cite{tibshirani1996regression}. Specifically, we assume
sparse coding is carried out by solving
\begin{align}
\label{eqn:lasso}
    x_{i,s} = \arg\min_{x \in \mathbb{R}^K} \tfrac{1}{2}{\|y_{i,s}-\Dictionary x \|_{2}^{2}} + \tau \|x\|_1
\end{align}
with the regularization parameter $\tau > 0$ selected in a way that
$\|x_{i,s}\|_0 \leq T_0 \ll n$. This can be accomplished, for example, by
making use of the \emph{least angle regression} algorithm
\cite{Efron.etal.AS2004}. Note that the lasso also has a dual, constrained
form, given by
\begin{align}
\label{eqn:dual_lasso}
    x_{i,s} = \arg\min_{x \in \mathbb{R}^K} \tfrac{1}{2}{\|y_{i,s}-\Dictionary x \|_{2}^{2}} \quad \text{s.t.} \quad \|x\|_1 \leq \eta,
\end{align}
and \eqref{eqn:lasso} \& \eqref{eqn:dual_lasso} are identical for an
appropriate $\eta_\tau = \eta(\tau)$ \cite{Figueiredo.etal.IJSTSP2007}.

{
\begin{remark}
While extension of our analysis to other sparse coding methods such as
\emph{orthogonal matching pursuit} (OMP)~\cite{tropp2007signal} is beyond the
scope of this work, such extensions would mainly rely on perturbation
analyses of different sparse coding methods. In the case of OMP, for
instance, such perturbation analysis is given in~\cite{ding2013perturbation},
which can then be leveraged to extend our lasso-based cloud K-SVD result to
OMP-based result.
%
\end{remark}
}

Our analysis in the following is also based on the assumption that cloud
K-SVD and centralized K-SVD are identically initialized, i.e.,
$\widehat{\Dictionary}_{i}^{(0)} = \Dictionary^{(0)}, i=1,\dots,N$, where
$\Dictionary^{(t)}, t \geq 0$, in the following denotes the centralized K-SVD
dictionary estimate in the $t^{th}$ iteration. While both cloud K-SVD and
centralized K-SVD start from the same initial estimates, the cloud K-SVD
dictionaries get perturbed in each iteration due to imperfect power method
and consensus averaging. In order to ensure these perturbations do not cause
the cloud K-SVD dictionaries to diverge from the centralized solution after
$T_d$ iterations, we need the dictionary estimates returned by centralized
K-SVD in each iteration to satisfy certain properties. Below, we present and
motivate these properties.
\begin{enumerate}
\item[\textbf{[P1]}] Let $x_{i,s}^{(t)}$ denote the solution of the lasso
    (i.e., \eqref{eqn:lasso}) for $D = D^{(t-1)}$ and $\tau = \tau^{(t)},
    t=1,\dots,T_d$. Then there exists some $C_1
    > 0$ such that the following holds:
    \begin{align*}
    \min\limits_{t,i,s,j\not\in\textrm{supp}(x_{i,s}^{(t)})} \tau^{(t)} - \big|\langle d_j^{(t)}, y_{i,s} - D^{(t-1)}x_{i,s}^{(t)}\rangle\big| > C_1.
    \end{align*}
    In our analysis in the following, we will also make use of the smallest
    regularization parameter among the collection
    $\big\{\tau^{(t)}\big\}_{t=1}^{T_d}$, defined as $\tau_{\min} =
    \min\limits_{t} \tau^{(t)}$, and the largest dual parameter among the
    (dual) collection $\big\{\eta_\tau^{(t)} = \eta(\tau^{(t)})
    \big\}_{t=1}^{T_d}$, defined as $\eta_{\tau,\max} = \max\limits_{t}
    \eta_\tau^{(t)}$.


\item[\textbf{[P2]}] Define $\Sigma_{T_0} = \big\{\mathcal{I} \subset
    \{1,\dots,K\} :  |\mathcal{I}| = T_0\big\}$. Then there exists some
    $C_2^{\prime} > \frac{C_1^4 \tau_{\min}^2}{1936}$ such that the
    following holds:
    \begin{align*}
        \min_{t=1,\dots,T_d, \mathcal{I} \in \Sigma_{T_0}} \sigma_{T_0}\left(D^{(t-1)}_{|\mathcal{I}}\right) \geq \sqrt{C_2^{\prime}},
    \end{align*}
    where $\sigma_{T_0}(\cdot)$ denotes the $T_0^{th}$ (ordered) singular
    value of a matrix. In our analysis, we will be using the parameter
    $C_2=\left(\sqrt{C_2^{\prime}}-\frac{C_1^2 \tau_{\min}}{44}\right)^2$.

\item[\textbf{[P3]}] Let $\lambda_{1,k}^{(t)} > \lambda_{2,k}^{(t)} \geq
    \dots \lambda_{n,k}^{(t)} \geq 0$ denote the eigenvalues of the
    centralized ``reduced'' matrix $E_{k,R}^{(t)}E_{k,R}^{(t)^{\tT}},
    k\in\{1,\dots,K\}$, in the $t^{th}$ iteration, $t\in\{1,\dots,T_d\}$. Then there
    exists some $C_3^{\prime} < 1$ such that the following holds:
    \begin{align*}
        \max_{t,k} \frac{\lambda_{2,k}^{(t)} }{\lambda_{1,k}^{(t)}} \leq C_3^{\prime}.
    \end{align*}
    Now define $C_3 = \max\Big\{1,
    \frac{1}{\min_{t,k}\lambda_{1,k}^{(t)}(1-C_3^{\prime})}\Big\}$, which we will use in our
    forthcoming analysis.
\end{enumerate}

We now comment on the rationale behind these three properties. Properties P1
and P2 correspond to sufficient conditions for $x_{i,s}^{(t)}$ to be a unique
solution of \eqref{eqn:lasso} \cite{Fuchs.ITIT2004} and guarantee that the
centralized K-SVD generates a unique collection of sparse codes in each
dictionary learning iteration. Property P3, on the other hand,
ensures that algorithms such as the power method can be used to compute the
dominant eigenvector of $E_{k,R}^{(t)}E_{k,R}^{(t)^{\tT}}$ in each dictionary
learning iteration \cite{golub2012matrix}. In particular, P3 is a statement
about the worst-case spectral gap of $E_{k,R}^{(t)}E_{k,R}^{(t)^{\tT}}$. In addition to these properties, our final analytical result for
cloud K-SVD will also be a function of a certain parameter of the centralized
error matrices $\big\{E_k^{(t)}\big\}_{k=1}^K$ generated by the centralized
K-SVD in each iteration. We define this parameter in the following for later
use. Let $E_{i,k}^{(t)}, i=1,\dots,N$, denote part of the centralized error
matrix $E_k^{(t)}$ associated with the data of the $i^{th}$ site in the
$t^{th}$ iteration, i.e., $E_k^{(t)} = \begin{bmatrix} E_{1,k}^{(t)} & \cdots & E_{N,k}^{(t)}\end{bmatrix}, k=1,\dots,K,
t=1,\dots,T_d$. Then
\begin{align}
\label{eqn:C_4_defn}
    C_4 = \max\Big\{1,\max_{t,i,k} \|E_{i,k}^{(t)}\|_2\Big\}.
\end{align}
\subsection{Main Result}
We are now ready to state the main result of this paper. This result is given
in terms of the $\|\cdot\|_2$ norm mixing time, $T_{mix}$, of the Markov
chain associated with the doubly-stochastic weight matrix $W$, defined as
\begin{align}
\label{eqn:Tmix}
    T_{mix} = \max_{i=1,\dots,N} \inf_{t \in \mathbb{N}}\left\{t : \|\mathrm{e}_i^\tT W^t - \tfrac{1}{N}\mathbf{1}^\tT\|_2 \leq \frac{1}{2}\right\}.
\end{align}
Here, $\mathrm{e}_i \in \mathbb{R}^N$ denotes the $i^{th}$ column of the
identity matrix $I_N$. Note that the mixing time $T_{mix}$ can be upper
bounded in terms of inverse of the absolute spectral gap of $W$, defined as
$1-|\lambda_2(W)|$ with $\lambda_2(W)$ denoting the second largest (in
modulus) eigenvalue of $W$ \cite{levin2009markov}. As a general rule,
better-connected networks can be made to have smaller mixing times compared
to sparsely connected networks. {We refer the reader
to~\cite{boyd2005mixing} and \cite[Chap.~15]{levin2009markov} for further
details on the relationship between $T_{mix}$ and the underlying network
topology.}

\begin{theorem}[Stability of Cloud K-SVD Dictionaries]\label{thm:main_result}
Suppose cloud K-SVD (Algorithm~\ref{algo:Distributed_KSVD}) and (centralized)
K-SVD are identically initialized and both of them carry out $T_d$ dictionary
learning iterations. In addition, assume cloud K-SVD carries out $T_p$ power
method iterations during the update of each atom and $T_c$ consensus
iterations during each power method iteration. Finally, assume the K-SVD
algorithm satisfies properties P1--P3. Next, define $\alpha = \max_{t,k}
\sum_{i=1}^N \|\widehat{E}_{i,k,R}^{(t)}\widehat{E}_{i,k,R}^{(t)^{\tT}}\|_2$,
$\beta = \max_{t,t_p,k}
\frac{1}{\left\|\widehat{E}_{k,R}^{(t)}\widehat{E}_{k,R}^{(t)^{\tT}}
q_{\textsf{c},t,k}^{(t_p)}\right\|_2}$, $\gamma =
\max_{t,k}\sqrt{\sum_{i=1}^{N}
\|\widehat{E}_{i,k,R}^{(t)}\widehat{E}_{i,k,R}^{(t)^{\tT}}\|_{F}^{2}}$, $\nu=\max_{t,k}{\frac{\widehat{\lambda}_{2,k}^{(t)}}{\widehat{\lambda}_{1,k}^{(t)}}}$, $\widehat{\theta}_k^{(t)}\in [0,\pi/2]$ as
    $\widehat{\theta}_k^{(t)}=\arccos{\left(\frac{\left|\left<u_{1,k}^{(t)},
    q^{init}\right>\right|}{\|u_{1,k}^{(t)}\|_2 \|q^{init}\|_2}\right)}$, $\mu =\max\big\{1,\max_{k,t}\tan(\widehat{\theta}_k^{(t)})\big\}$, and
$\zeta = K \sqrt{2S_{\max}} \left(\frac{6\sqrt{K T_0}}{\tau_{\min}
C_2}+\eta_{\tau,\max}\right)$, where $S_{\max} = \max_{i} S_i$, $u_{1,k}^{(t)}$ is the dominant eigenvector of $\widehat{E}_{k,R}^{(t)}\widehat{E}_{k,R}^{(t)^{\tT}}$, $\widehat{\lambda}_{1,k}^{(t)}$ and $\widehat{\lambda}_{2,k}^{(t)}$ are first and second largest eigenvalues of $\widehat{E}_{k,R}^{(t)}\widehat{E}_{k,R}^{(t)^{\tT}}$ respectively, and
$q_{\textsf{c},t,k}^{(t_p)}$ denotes the iterates of a centralized power
method initialized with $q^{init}$ for estimation of the dominant eigenvector
of $\widehat{E}_{k,R}^{(t)}\widehat{E}_{k,R}^{(t)^{\tT}}$. Then, assuming $\min_{t,k} \big|\langle u_{1,k}^{(t)}, q^{init}\rangle\big| > 0$, and fixing any $\epsilon \in \Big(0,\min\big\{(10\alpha^2\beta^2)^{-1/3T_p},(\frac{1-\nu}{4})^{1/3}\big\}\Big)$
and $\delta_d \in \Big(0,\min\big\{\frac{1}{\sqrt{2}}, \frac{C_1^2
\tau_{\min}}{44\sqrt{2K}}\big\}\Big)$, we have
\begin{align}
\label{enq:thm_eqn_main}
    \max_{\substack{i=1,\dots,N\\k=1,\dots,K}} \left\|\widehat{d}_{i,k}^{(T_d)}\widehat{d}_{i,k}^{(T_d)^{\tT}} - d_k^{(T_d)}d_k^{(T_d)^{\tT}}\right\|_2 \leq \delta_d
\end{align}
as long as the number of power method iterations $T_p \geq \frac{2(T_d K -
2)\log(8C_3 C_4^2 N+5) + (T_d -1)\log(1+\zeta) + \log(8 C_3 C_4 \mu
N\sqrt{n}\delta_d^{-1})}{\log[(\nu + 4\epsilon^3)^{-1}]}$ and the number of
consensus iterations $T_c = \Omega\big(T_p T_{mix} \log{(2\alpha \beta
\epsilon^{-1})}+T_{mix} \log{(\alpha^{-1}\gamma\sqrt{N})}\big)$.
\end{theorem}

The proof of this theorem is given in Appendix~\ref{app:proof_mainthm}. We
now comment on the major implications of Theorem~\ref{thm:main_result}.
First, the theorem establishes that the distributed dictionaries
$\{\widehat{D}_{i}^{(T_d)}\}$ can indeed remain arbitrarily close to the
centralized dictionary $D^{(T_d)}$ after $T_d$ dictionary learning iterations
(cf.~\ref{enq:thm_eqn_main}). Second, the theorem shows that this can happen
as long as the number of distributed power method iterations $T_p$ scale in a
certain manner. In particular, Theorem~\ref{thm:main_result} calls for this
scaling to be at least linear in $T_d K$ (modulo the $\log{N}$ multiplication
factor), which is the total number of SVDs that K-SVD needs to perform in
$T_d$ dictionary learning iterations. On the other hand, $T_p$ need only
scale logarithmically with $S_{\max}$, which is significant in the context of
big data problems. Other main problem parameters that affect the scaling of
$T_p$ include $T_0$, $n$, and $\delta_d^{-1}$, all of which enter the scaling
relation in a logarithmic fashion. Finally, Theorem~\ref{thm:main_result}
dictates that the number of consensus iterations $T_c$ should also scale at
least linearly with $T_p T_{mix}$ (modulo some $\log$ factors) for the main
result to hold. {Notice that the effect of network topology on the
number of consensus iterations is captured through the dependence of $T_c$ on
the mixing time $T_{mix}$.} In summary, Theorem~\ref{thm:main_result}
guarantees that the distributed dictionaries learned by cloud K-SVD can
remain close to the centralized dictionary without requiring excessive
numbers of power method and consensus averaging iterations.

We now provide a brief heuristic understanding of the roadmap needed to prove
Theorem~\ref{thm:main_result}. In the first dictionary learning iteration
($t=1$), we have $\{\widehat{D}_{i}^{(t-1)} \equiv D^{(t-1)}\}$ due to
identical initializations. While this means both K-SVD and cloud K-SVD result
in identical sparse codes for $t=1$, the distributed dictionaries begin to
deviate from the centralized dictionary after this step. The perturbations in
$\{\widehat{d}_{i,k}^{(1)}\}$ happen due to the finite numbers of power
method and consensus averaging iterations for $k=1$, whereas they happen for
$k > 1$ due to this reason as well as due to the earlier perturbations in
$\{\widehat{d}_{i,j}^{(1)}, \widehat{x}_{i,j,T}^{(1)}\}, j < k$. In
subsequent dictionary learning iterations ($t>1$), therefore, cloud K-SVD
starts with already perturbed distributed dictionaries
$\{\widehat{D}_{i}^{(t-1)}\}$. This in turn also results in deviations of the
sparse codes computed by K-SVD and cloud K-SVD, which then adds another
source of perturbations in $\{\widehat{d}_{i,k}^{(t)}\}$ during the
dictionary update steps. To summarize, imperfect power method and consensus
averaging in cloud K-SVD introduce errors in the top eigenvector estimates of
(centralized) $E_{1,R}^{(1)} E_{1,R}^{(1)^{\tT}}$ at individual sites, which
then accumulate for $(k,t) \not= (1,1)$ to also cause errors in estimate
$\widehat{E}_{k,R}^{(t)} \widehat{E}_{k,R}^{(t)^{\tT}}$ of the matrix
$E_{k,R}^{(t)} E_{k,R}^{(t)^{\tT}}$ available to cloud K-SVD. Collectively,
these two sources of errors cause deviations of the distributed dictionaries
from the centralized dictionary and the proof of
Theorem~\ref{thm:main_result} mainly relies on our ability to control these
two sources of errors.

\subsection{Roadmap to Theorem~\ref{thm:main_result}}
The first main result needed for the proof of Theorem~\ref{thm:main_result}
looks at the errors in the estimates of the dominant eigenvector $u_1$ of an
arbitrary symmetric matrix $M = \sum_{i=1}^N M_i$ obtained at individual
sites using imperfect power method and consensus averaging when the $M_i$'s
are distributed across the $N$ sites
(cf.~Sec.~\ref{subsec:Distributed_K-SVD}). The following result effectively
helps us control the errors in cloud K-SVD dictionaries due to Steps 7--17 in
Algorithm~\ref{algo:Distributed_KSVD}.
\begin{theorem}[Stability of Distributed Power Method]
\label{theorem:DistributedPowerMethod} Consider any symmetric matrix $M =
\sum_{i=1}^{N} M_i$ with dominant eigenvector $u_1$ and eigenvalues
$|\lambda_1| > |\lambda_2| \geq \cdots \geq |\lambda_n|$. Suppose each $M_i,
i=1,\dots,N$, is only available at the $i^{th}$ site in our network and let
$\widehat{q}_i$ denote an estimate of $u_1$ obtained at site $i$ after $T_p$
iterations of the distributed power method (Steps 7--17 in
Algorithm~\ref{algo:Distributed_KSVD}). Next, define $\alpha_p = \sum_{i=1}^N
\|M_i\|_2$, $\beta_p = \max_{t_p = 1,\dots,T_p} \frac{1}{\|M
q_{\textsf{c}}^{(t_p)}\|_2}$, and $\gamma_p =
\sqrt{\sum_{i=1}^{N}{\|M_i\|_{F}^{2}}}$, where $q_{\textsf{c}}^{(t_p)}$
denotes the iterates of a centralized power method initialized with
$q^{init}$. Then, fixing any $\epsilon \in \big(0,
(10\alpha_p^2\beta_p^2)^{-1/3T_p}\big)$, we have
\begin{align}
\label{eq:mainBound}
\max_{i=1,\dots,N} \left\|u_1 u_1^\tT - \widehat{q}_i\widehat{q}_i^\tT\right\|_2 \leq \tan{(\theta)}
\left|\frac{\lambda_{2}}{\lambda_{1}}\right|^{T_p}+ 4\epsilon^{3T_p},
\end{align}
as long as $|\langle u_1, q^{init}\rangle| >0$ and the number of consensus
iterations within each iteration of the distributed power method (Steps
10--14 in Algorithm~\ref{algo:Distributed_KSVD}) satisfies $T_c =
\Omega\big(T_p T_{mix} \log{(2\alpha_p \beta_p \epsilon^{-1})}+T_{mix}
\log{(\alpha_p^{-1}\gamma_p\sqrt{N})}\big)$. Here, $\theta$ denotes the angle
between $u_1$ and $q^{init}$, defined as $\theta = \arccos(|\langle u_1,
q^{init}\rangle|/(\|u_1\|_2\|q^{init}\|_2))$.
\end{theorem}

The proof of this theorem is given in Appendix~\ref{app:proof_DPM}.
Theorem~\ref{theorem:DistributedPowerMethod} states that $\widehat{q}_i
\stackrel{T_p}{\longrightarrow} \pm u_1$ geometrically at each site
as long as enough consensus iterations are performed in each iteration of the
distributed power method. In the case of a finite number of distributed power
method iterations, \eqref{eq:mainBound} in
Theorem~\ref{theorem:DistributedPowerMethod} tells us that the maximum error
in estimates of the dominant eigenvector is bounded by the sum of two terms,
with the first term due to finite number of power method iterations and the
second term due to finite number of consensus iterations.

The second main result needed to prove Theorem~\ref{thm:main_result} looks at the errors between individual blocks of the reduced distributed error matrix $\widehat{E}_{k,R}^{(t)}=\begin{bmatrix}\widehat{E}_{1,k,R}^{(t)},\cdots,\widehat{E}_{N,k,R}^{(t)} \end{bmatrix}$ and the reduced centralized error matrix $E_{k,R}^{(t)}=\begin{bmatrix}E_{1,k,R}^{(t)},\cdots,E_{N,k,R}^{(t)}\end{bmatrix}$ for $k\in \{1,\cdots,K\}$ and $t\in\{1,\cdots,T_d\}$. This result helps us control the error in step~5 of Algorithm~\ref{algo:Distributed_KSVD} and, together with Theorem~\ref{theorem:DistributedPowerMethod}, characterizes the major sources of errors in cloud $K$-SVD in relation to centralized $K$-SVD. The following theorem provides a bound on error in $E_{i,k,R}^{(t)}$

\begin{theorem}[Perturbation in the matrix $\widehat{E}_{i,k,R}^{(t)}$]
\label{thm:AllIteration} Recall the definitions of $\Omega_{k}^{(t)}$ and
$\widetilde{\Omega}_{i,k}^{(t)}$ from Sec.~\ref{subsec:reviewKSVD} and
Sec.~\ref{subsec:Distributed_K-SVD}, respectively. Next, express
$\Omega_k^{(t)}=\text{diag}\{\Omega_{1,k}^{(t)},\cdots,
\Omega_{N,k}^{(t)}\}$, where $\Omega_{i,k}^{(t)}$ corresponds to
the data samples associated with the $i^{th}$ site, and define
$B_{i,k,R}^{(t)}=\widehat{E}_{i,k,R}^{(t)}-E_{i,k,R}^{(t)}$. Finally, let
$\zeta$, $\mu$, $\nu$, $\epsilon$, and $\delta_d$ be as in
Theorem~\ref{thm:main_result}, define $\varepsilon=\mu
\nu^{T_p}+4\epsilon^{3T_p}$, and assume $\varepsilon \leq
\frac{\delta_d}{8N\sqrt{n} C_3 (1+\zeta)^{T_d-1} C_4^{2} (8C_3 N
C_4^{2}+5)^{2(T_d K-2)}}$. Then, if we perform $T_p$ power method iterations
and $T_c= \Omega\big(T_p T_{mix} \log{(2\alpha \beta \epsilon^{-1})}+T_{mix}
\log{(\alpha^{-1}\gamma\sqrt{N})}\big)$ consensus iterations in cloud $K$-SVD
and assume P1--P3 hold, we have for $i\in\{1,\dots,N\}$,
$t\in\{1,\cdots,T_d\}$, and $k\in\{1,\cdots,K\}$
\begin{align*}
\|B_{i,k,R}^{(t)}\|_2 \leq
\begin{cases}\label{eq:Bound_B}
0,\qquad\text{for } t=1,k=1,\\
\varepsilon (1+\zeta)^{t-1}C_4 (8C_3 N C_4^2+5)^{(t-1)K+k-2},\ \text{o.w}.
\end{cases}
\end{align*}
\end{theorem}
Proof of Theorem~\ref{thm:AllIteration} along with the proofs of supporting
lemmas is given in Appendix~\ref{app:proof_thm_Bik}.
Theorem~\ref{thm:AllIteration} tells us that the error in matrix
$E_{i,k,R}^{(t)}$ can be made arbitrarily small through a suitable choice of
$T_p$ and $\epsilon$ as long as all of the assumptions of
Theorem~\ref{thm:main_result} are satisfied. The proof
of Theorem~\ref{thm:main_result}, given in Appendix~\ref{app:proof_mainthm},
relies on these two aforementioned theorems. In particular, the proof of
Theorem~\ref{thm:main_result} shows that the assumption on $\varepsilon$
in Theorem~\ref{thm:AllIteration} is satisfied as long as we are performing power method iterations and
consensus iterations as required by Theorem~\ref{thm:main_result}.

%% file: docs/NumericalEvaluationV2.tex
\begin{figure*}[t]
\centering
\subfigure[Distributed Power Method]{\includegraphics[width=1.6in]{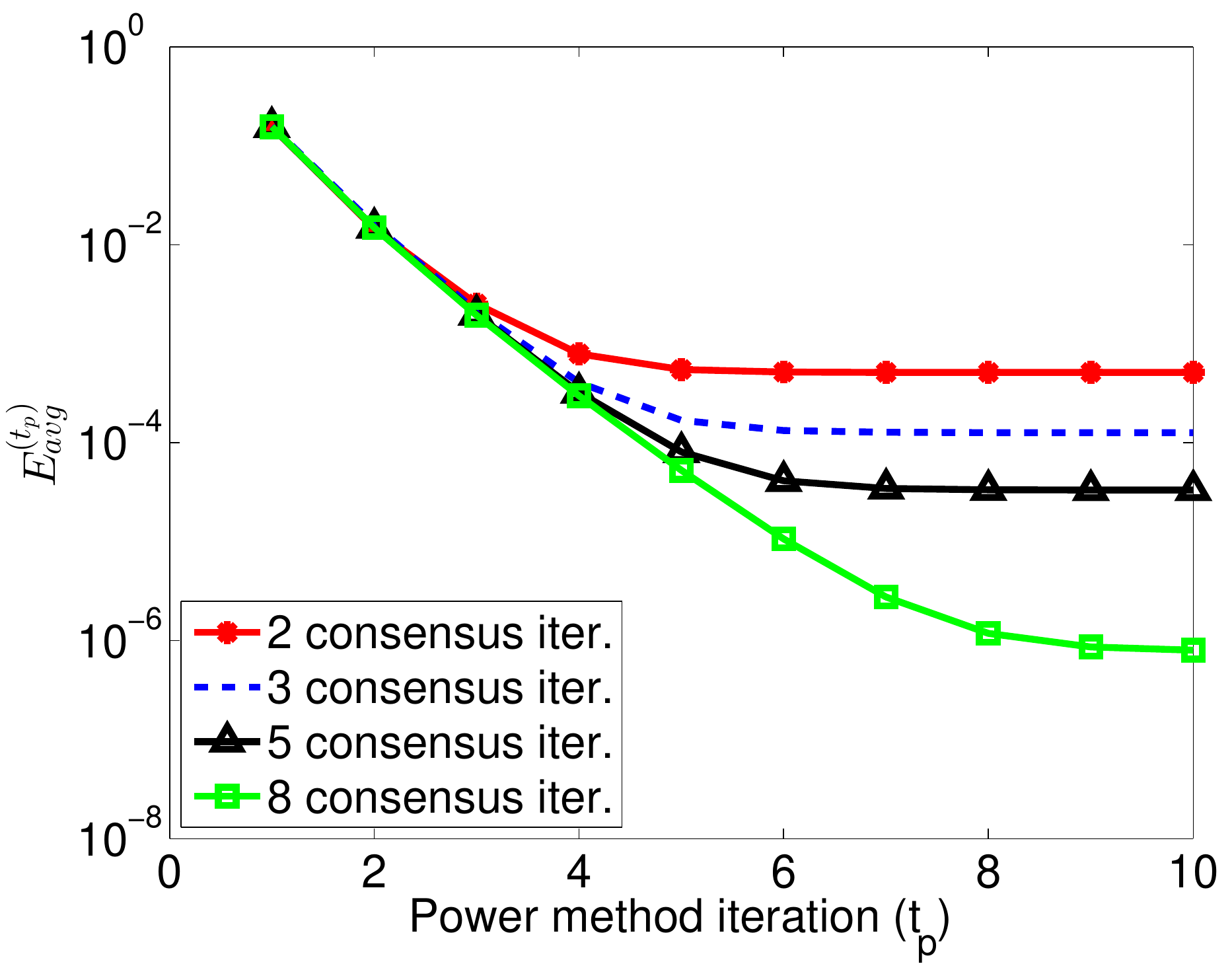} \label{fig:DistributedPowerMethod}}
\subfigure[Cloud K-SVD Error]{\includegraphics[width=1.6in]{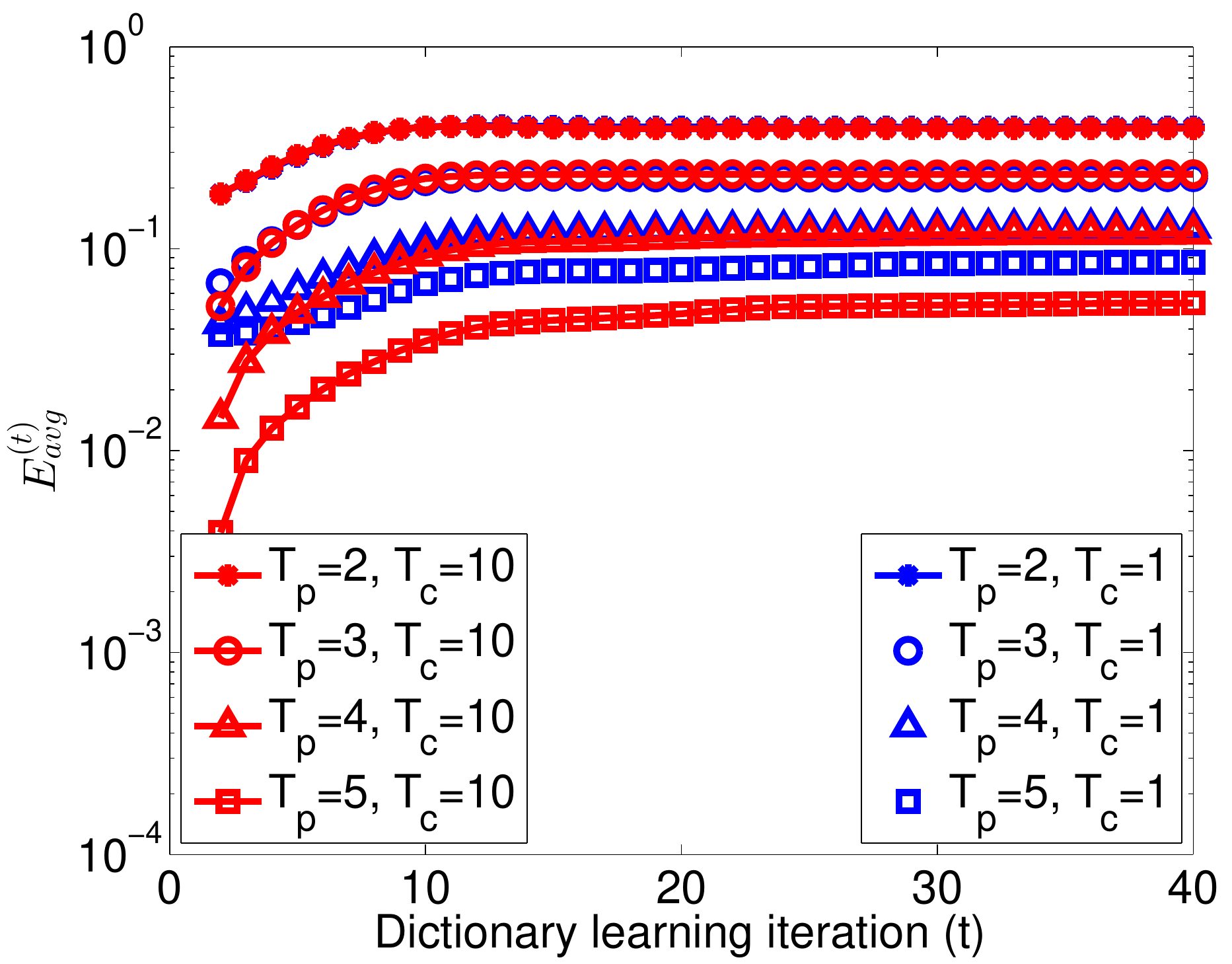} \label{fig:DictionaryErrorVaryingPowerIterations}}
\subfigure[Average Representation Error]{\includegraphics[width=1.6in]{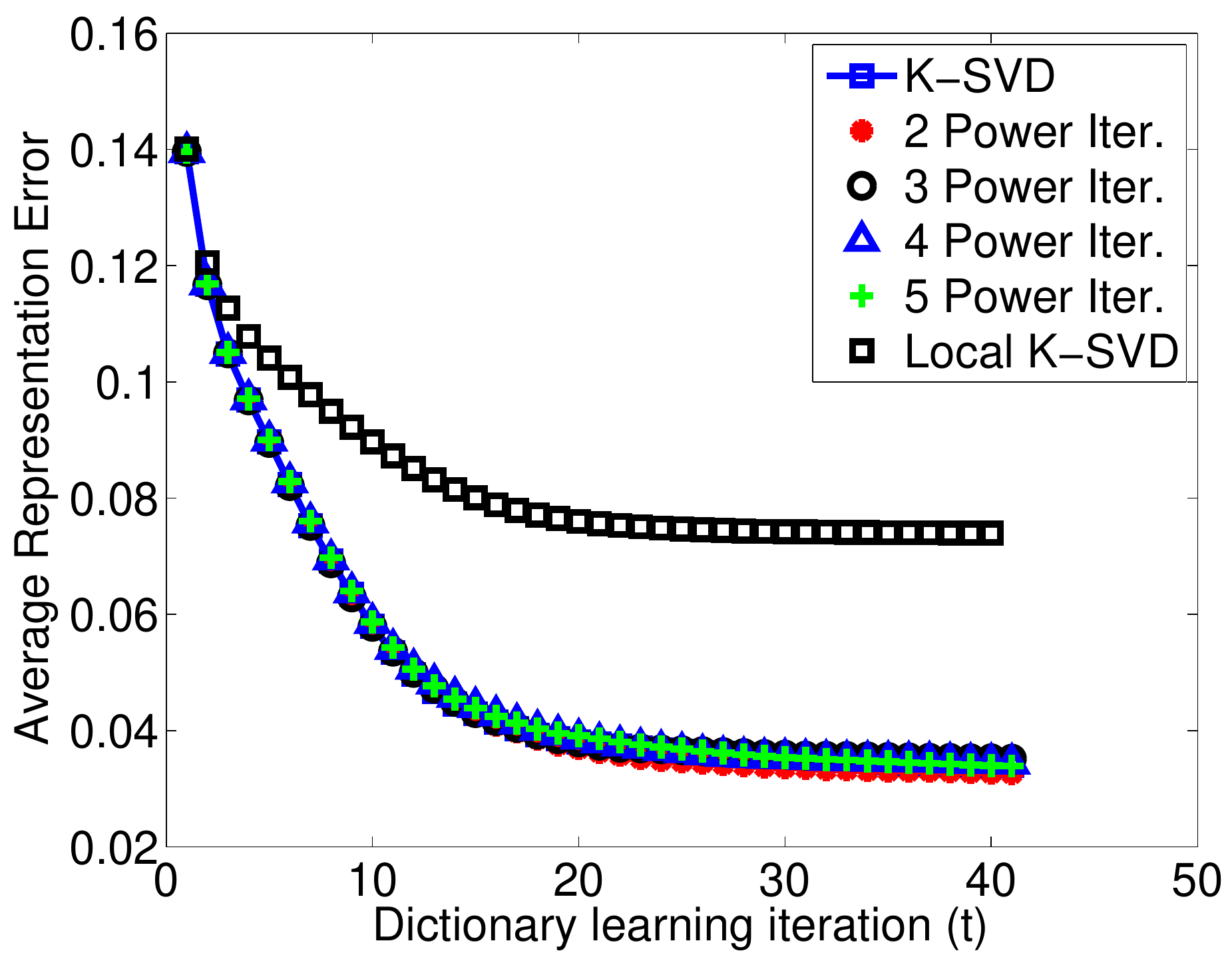} \label{fig:RepresentationErrorVaryingPowerIterations}}
\subfigure[Online K-SVD]{\includegraphics[width=1.6in]{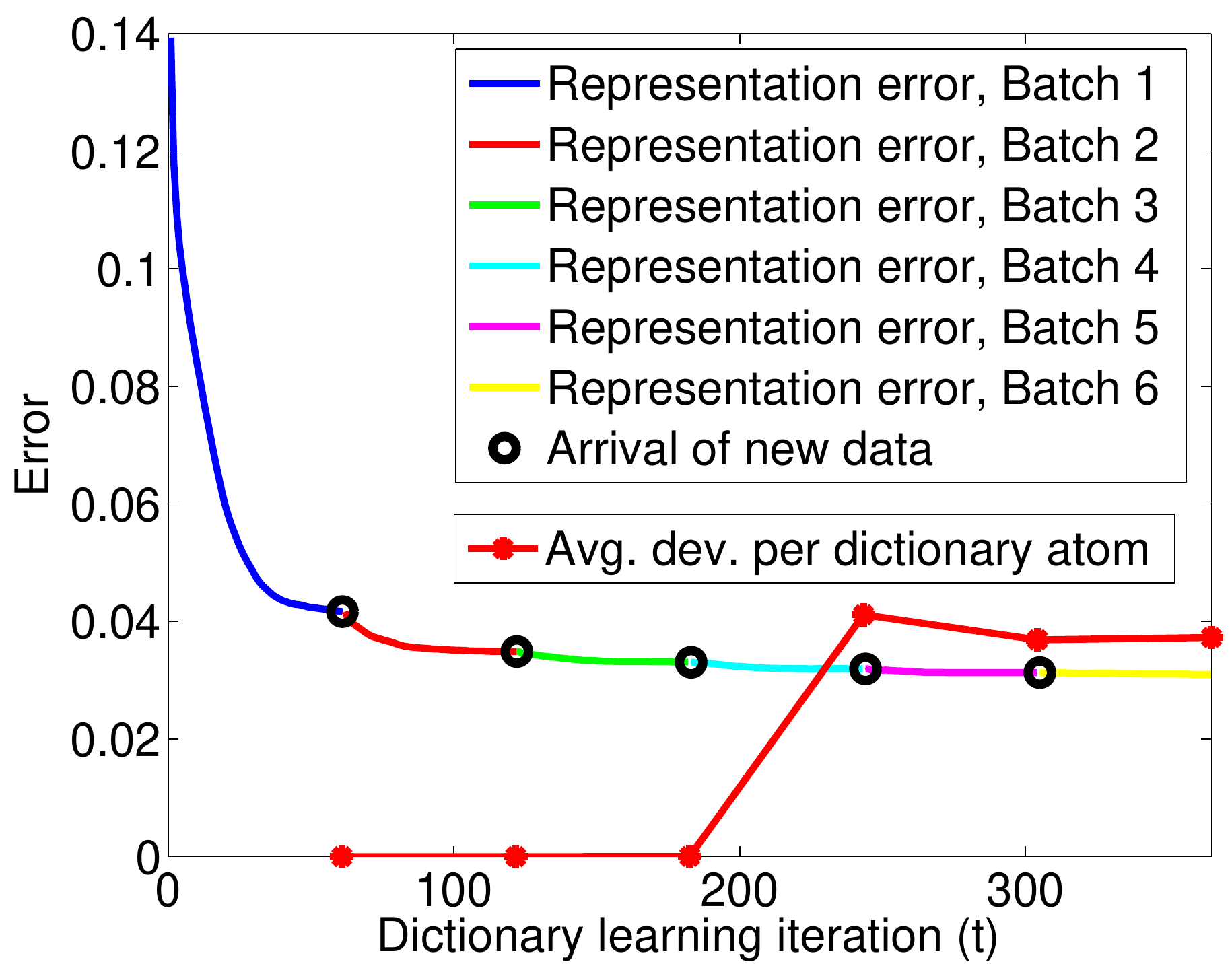} \label{fig:OnlineKSVD}}
\caption{Performance of cloud K-SVD on synthetic data.
(a) Average error in eigenvector estimates of distributed power method. (b) Average error in
dictionary atoms returned by cloud K-SVD. (c) Average representation error of cloud K-SVD.
(d) Average representation error \revtwo{and average deviation per dictionary atom (from centralized dictionary learned in a full-batch setting)} of K-SVD in an online setting as a function of dictionary learning iterations.}
\label{fig:numerical_results}
\vspace{-1.5\baselineskip}
\end{figure*}

\section{Numerical Experiments}\label{section:NumericalEvaluation}
We present numerical results in this section for demonstrating
the usefulness of cloud K-SVD and also validating some of our theoretical results. In the first set of experiments, synthetic data is used to demonstrate efficacy of cloud $K$-SVD for data representation. Furthermore, behavior of distributed power method (Steps~7--17 in Algorithm~\ref{algo:Distributed_KSVD}) as a function of the number of consensus iterations and deviations in cloud $K$-SVD dictionaries from centralized dictionary as a function of number of power method iterations are also shown with the help of simulations. In the second set of experiments, MNIST dataset is used to motivate an application of cloud $K$-SVD that can benefit from collaboration between distributed sites.
%
\subsection{Experiments Using Synthetic Data}
\label{subsec:syntheticData} These experiments correspond to a total of $N =
100$ sites, with each site having $S_i = 500$ local samples in
$\mathbb{R}^{20}$ (i.e., $n = 20$). Interconnections between the sites are
randomly generated using an Erd\H{o}s--R\'{e}nyi graph with parameter
$p=0.5$. In order to generate synthetic data at individual sites, we first
generate a dictionary with $K=50$ atoms, $\Dictionary \in \mathbb{R}^{20
\times 50}$, with columns uniformly distributed on the unit sphere in
$\mathbb{R}^{20}$. Next, we randomly select a $45$-column subdictionary of
$\Dictionary$ for each site and then generate samples for that site using a
linear combination of $T_0 = 3$ randomly selected atoms of this
subdictionary, followed by addition of white Gaussian noise with variance
$\sigma^2 = 0.01$. All data samples in our experiments are also normalized to
have unit $\ell_2$ norms. Sparse coding in these experiments is performed
using an implementation of OMP provided in~\cite{rubinstein2008efficient}.
Finally, in order to carry out distributed consensus averaging, we generate a
doubly-stochastic weight matrix $W$ according to the local-degree weights
method described in \cite[Sec.~4.2]{xiao2004fast}.

In our first set of experiments we illustrate the convergence behavior of the
distributed power method component within cloud K-SVD (Steps 7--17 in
Algorithm~\ref{algo:Distributed_KSVD}) as a function of the number of
consensus iterations. The results of these experiments, which are reported in
Fig.~\ref{fig:DistributedPowerMethod}, correspond to five different values of
the number of consensus iterations (3, 4, 5, 10, 15) within each iteration of
the distributed power method. Specifically, let $q$ denote the principal
eigenvector of the matrix $\sum_{i=1}^{N}{\widehat{M}_i}$ in
Algorithm~\ref{algo:Distributed_KSVD} (Step 6) computed using Matlab
(ver.~2014a) and $\widehat{q}_i^{(t_p)}$ denote an estimate of $q$ obtained
at site $i$ after the $t_p^{th}$ iteration of the distributed power method.
Then Fig.~\ref{fig:DistributedPowerMethod} plots $E_{\text{eig}}^{(t_p)}$,
which is the average of ${\|q q^\tT -
\widehat{q}_i^{(t_p)}{\widehat{q}_i^{(t_p) \tT}}\|_2}$ over all sites $i\in
\{1,\dots, N\}$, dictionary update steps $k\in\{1,\dots,K\}$, dictionary
learning iterations $T_d$, and 100 Monte-Carlo trials, as a function of the
number of distributed power method iterations $t_p$. It can be seen from this
figure that the distributed power method of
Algorithm~\ref{algo:Distributed_KSVD} hits an \emph{error floor} with
increasing number of distributed power method iterations, where the floor is
fundamentally determined by the number of consensus iterations within each
power method iteration, as predicted by
Theorem~\ref{theorem:DistributedPowerMethod}.

Using the same setup our second set of experiments demonstrate the
effectiveness of collaboratively learning a dictionary using cloud K-SVD, as
opposed to each site learning a \emph{local dictionary} from its local data
using the canonical K-SVD algorithm (referred to as \emph{local K-SVD} in the
following). Moreover, these experiments also demonstrate the variations in
cloud K-SVD results when we change the number of power method iterations ($T_p$) and
consensus iterations ($T_c$). In
Fig.~\ref{fig:RepresentationErrorVaryingPowerIterations}, we plot average
representation error, defined as $\frac{1}{n
S}\sum_{i=1}^{N}{\sum_{j=1}^{S_i}{\|y_{i,j} - D x_{i,j}\|_2}}$, as a function
of the number of dictionary learning iterations for three dictionary learning
methods, namely, centralized (canonical) K-SVD, cloud K-SVD, and local K-SVD.
It can be seen from this figure, which corresponds to an average of 100
Monte-Carlo trials, that cloud K-SVD and centralized $K$-SVD have similar
performance and both of them perform better than local K-SVD. In particular,
the local K-SVD error is $\approx 0.06$ after 40 iterations, while it is
$\approx 0.03$ for cloud K-SVD and centralized K-SVD. {Notice that
changes in the number of power method iterations induce relatively minor
changes in the representation error of cloud K-SVD. Next,
Fig.~\ref{fig:DictionaryErrorVaryingPowerIterations} highlights the average
error in dictionary atoms learned using cloud K-SVD as compared to
centralized K-SVD. For this experiment, number of consensus iterations are
either $T_c=1$ or $T_c=10$, and for each of these values, the number of power
method iterations used are $T_p= 2, 3, 4, 5$. These experiments show the
effect of changing $T_p$ and $T_c$ on the error in collaborative dictionaries.} This error is averaged over
all dictionary atoms and sites in each iteration for 100 Monte-Carlo trials, defined as
$E_{\text{average}}^{(t)}=\frac{1}{NK}\sum_{k=1}^{K}\sum_{i=1}^{N}{\|d_k^{(t)}
d_k^{(t)^{\tT}} - \widehat{d}_{i,k}^{(t)}
\widehat{d}_{i,k}^{(t)^{\tT}}\|_2}$. Results in
Fig.~\ref{fig:DictionaryErrorVaryingPowerIterations} show that this error in
dictionary atoms increases sharply at the start, but it stabilizes after some
iterations. Important point to note here is that as we increase the number of
power method iterations {and consensus iterations} we get smaller
average error in dictionary atoms as predicted by our analysis.

Next, we discuss the usage of cloud K-SVD in online settings. Since it has
already been demonstrated that cloud K-SVD achieves performance similar to
that of K-SVD, we focus here on the representation error of centralized K-SVD
in online settings. The setup corresponds to a mini-batch of 500 training
samples being periodically generated at each site and the assumption that
each site has a buffer limit of 1000 samples. Thus only samples from the last
two periods can be used for dictionary learning. After arrival of each new
mini-batch of training samples, we use the dictionary learned in the last
period to warm-start (centralized) K-SVD and carry out 60 dictionary learning
iterations. Fig.~\ref{fig:OnlineKSVD} shows the representation error of the
learned dictionary in this case, along with the deviation per dictionary atom
when compared to a dictionary learned using full-batch centralized K-SVD.
These results are plotted as a function of dictionary learning iterations for
six periods, where the ending of a period is marked by a circle. The
representation error curve in this figure shows that K-SVD takes more time to
converge, but it (and thus cloud K-SVD) is a viable option for online
settings. Similarly, the deviation curve shows that while the dictionary
error initially increases with the arrival of more data, it stabilizes
afterward. Note that further improvements in these results can be obtained by
using methods like~\cite{tong2015active} for active sample selection.

{Finally, we perform experiments to report actual values of the
parameters $C_1$--$C_4$. To this end, we generate samples belonging to
$\mathbb{R}^{17}$, where each sample is a linear combination of $T_0=3$ atoms
of a dictionary $D\in\mathbb{R}^{17\times 40}$. We perform sparse coding in
these experiments using the lasso package in Matlab 2014a, while we perform
dictionary learning using K-SVD. Average values obtained for parameters
$C_1$--$C_4$ over 100 Monte-Carlo trials in this case are 0.0586, 0.1633,
4.544, and 1.5947, respectively. Using cloud K-SVD, average values of $\mu$
and $\nu$ are 9000 and 0.3242, respectively. Based on these values, we get
$T_p \approx 16,000$. This suggests that the constants in our bounds are
rather loose, and our analysis should mainly be used to provide scaling
guidelines.}

\subsection{{Classification of MNIST Images}}
\label{subsec:realData} For evaluation of cloud K-SVD on real dataset, we
perform classification of digits $\{0, 3, 5, 8, 9\}$ from MNIST
dataset~\cite{mnist-database}. For each digit~6000 samples are used,
where~5000 samples are used for training purposes and remaining~1000 for
testing purposes. The data are five-times randomly split into training and test
samples. For cloud $K$-SVD, Erd\H{o}s--R\'{e}nyi graph with parameter $p=0.5$
is used to generate a network with~10 sites and data is equally distributed
among them. Before performing dictionary learning, data is down sampled from
$\mathbb{R}^{784}$ to $\mathbb{R}^{256}$. After downsampling, a separate
dictionary is learned for each digit using centralized K-SVD, cloud K-SVD,
and K-SVD using only local data. Each dictionary has dimensions
$\mathbb{R}^{256\times 400}$, i.e., $K=400$, and sparsity level of $T_0=10$
is used. Minimum residue based rule~\cite[Sec.II-A]{chen2011hyperspectral} is
used for classification, more details on which are given in the following
paragraph.

\begin{figure}[t]
	\centering
		\includegraphics[width=0.5\columnwidth]{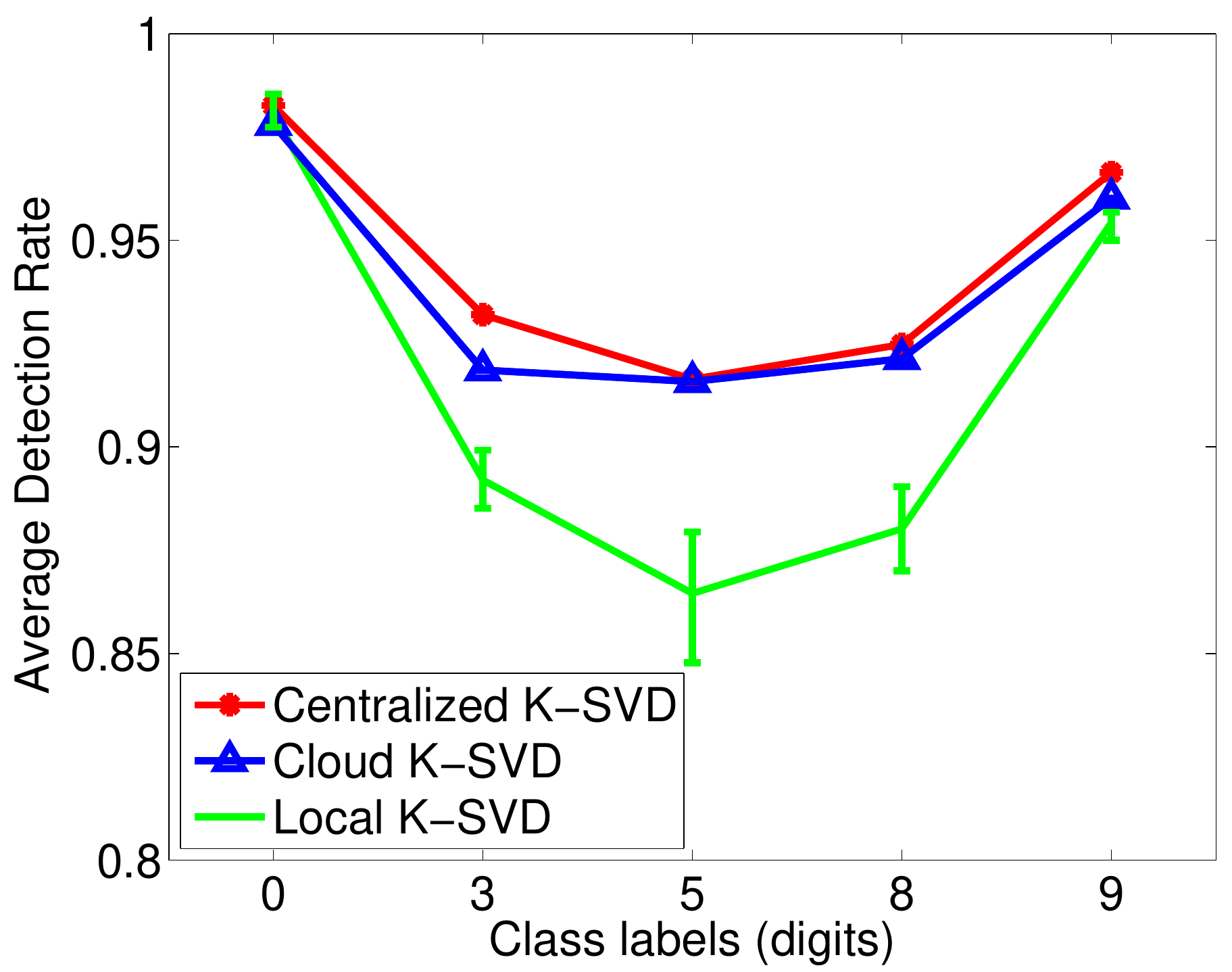}
	\caption{Average detection rate for five classes of MNIST dataset using centralized K-SVD, cloud K-SVD, and local K-SVD.}
	\vspace{-1.5\baselineskip}
	\label{fig:Classification_MNIST}
\end{figure}

Let $\{D_c\}_{c=1}^{5}$ be the set of dictionaries for~5 classes and let
$D=\begin{bmatrix} D_1 & D_2 & D_3 & D_4 & D_5\end{bmatrix}$ be the complete
dictionary. For any test sample $y_s$, we perform sparse coding using
dictionary $D$ with sparsity constraint of $T_0=10$ to get coefficients
$x_s\in \mathbb{R}^{2000}$. Then we partition $x_s$ into five segment
$\{x_{s,c}\}_{c=1}^{5}$, where $x_{s,c}$ are the coefficients corresponding
to dictionary $D_c$ of class $c$. Next we define residue for class $c$ as
$r_c = \|y_s - D_c x_{s,c}\|_2$. Finally, the detected class is given by
$c^{*}=\arg\min_{c}{r_c}$. Performance of each method (centralized $K$-SVD,
cloud $K$-SVD, and local $K$-SVD) is measured in terms of average detection
rate on the test samples, which is defined as $R_c = \frac{\text{Number of
samples in class }c\text{ detected correctly}}{\text{Total number of samples
of class } c}.$ Results of this experiment are given in
Fig.~\ref{fig:Classification_MNIST}. We see that centralized and cloud
$K$-SVD have comparable performance. But in the case of local $K$-SVD where
we only use the local data for learning representations, classification rate
deteriorates considerably. The bars in local $K$-SVD show the highest and
lowest detection rates achieved among the $10$~sites, which highlights the
variation in effectiveness of models learned across different sites when
using only the local data.

%% file: docs/Conclusion.tex
\section{Conclusion}
\label{section:conclusion} In this paper, we have proposed a new dictionary
learning algorithm, termed cloud K-SVD, that facilitates collaborative
learning of a dictionary that best approximates massive data distributed
across geographical regions. Mathematical analysis of proposed method is also
provided, which under certain assumptions shows that if we perform enough
 number of power method and consensus iterations then the proposed algorithm converges
to the centralized K-SVD solution. Furthermore, the efficacy of the proposed algorithm
is demonstrated through extensive simulations on synthetic and real data.
%

%% file: docs/Appendix.tex
\begin{appendices}

\section{Proof of Theorem~\ref{theorem:DistributedPowerMethod}}\label{app:proof_DPM}
The proof of this theorem relies on a lemma that guarantees that if the
estimates obtained at different sites using the distributed power method are
close to the estimate obtained using the centralized power method at the
start of a power method iteration then the distributed estimates remain close
to the centralized estimate at the end of that iteration. To prove such a
lemma, we first need a result from the literature that characterizes the
convergence behavior of \emph{vector} consensus averaging as a function of
the number of consensus iterations.
\begin{proposition}\cite[Theorem~5]{kempe2008decentralized}
\label{theorem:ConsensusError} Consider the $n \times 1$ vector sum $z =
\sum_{i=1}^N z_i^{(0)}$ and suppose each vector $z_i^{(0)}, i=1,\dots,N$, is
only available at the $i^{th}$ site in our network. Let $b$ be a vector whose entries are the sum of absolute values of the
initial vectors $z_i^{(0)}$ (i.e., $j^{th}$ entry of $b$ is $b_j=\sum_{i=1}^{N}{|z_{i,j}^{(0)}|}$) and $z_i^{(t_c)}$ be the $n \times 1$ vector
obtained at the $i^{th}$ site after $t_c$ consensus iterations. Then, fixing
any $\delta > 0$, we have that $\Big\|\frac{z_i^{(t_c)}}{[W^{t_c}_1]_i}-
z\Big\|_2\leq \delta \|b\|_{2}$ $\forall i$ as long as the number of
consensus iterations satisfies $t_c = \Omega(T_{mix}\log{\delta^{-1}})$.
\end{proposition}

We use Proposition~\ref{theorem:ConsensusError} to state and prove the
desired lemma.

\begin{lemma}
\label{le:ErrorInIteration} Suppose we are at the start of $(t_p + 1) \leq
T_p$ power method iteration. Let $q_{\textsf{c}}$ and $q_{i,\textsf{d}}$
denote the outputs of centralized power method and distributed power method
at $i^{th}$ site after $t_p$ iterations, respectively. Similarly, let
$q_{\textsf{c}}^\prime$ and $q_{i,\textsf{d}}^\prime$ denote the outputs of
centralized power method and distributed power method at $i^{th}$ site after
$t_p+1$ iterations, respectively. Next, fix an $\epsilon \in (0,1)$, define
$\delta = \frac{\alpha_p}{ \gamma_p
\sqrt{N}}\big(\frac{\epsilon}{2\alpha_p\beta_p}\big)^{3T_p}$, and assume that
$\forall i$, ${\|q_{\textsf{c}}-q_{i,\textsf{d}}\|_2} + \frac{\delta\gamma_p
\sqrt{N}}{\alpha_p}\leq \frac{1}{2\alpha_p\beta_p^2(2\alpha_p+\delta\gamma_p
\sqrt{N})}$. Then, assuming $\Omega(T_{mix}\log{\delta^{-1}})$ consensus
iterations, we have that%
\vspace{-\baselineskip}%
{\small$$\forall i, \
\|q_{\textsf{c}}^\prime-q_{i,\textsf{d}}^\prime\|_2 \leq (2\alpha_p
\beta_p)^{3}\left(\max_{i=1,\dots,N}\|q_{\textsf{c}}-q_{i,\textsf{d}}\|_2+\frac{\delta\gamma_p
\sqrt{N}}{\alpha_p}\right).$$}
\end{lemma}
\vspace{-0.35\baselineskip}%
\begin{IEEEproof} Define $v = M q_{\textsf{c}}$ and $\widehat{v} =
\sum_{i=1}^{N}{M_{i}q_{i,\textsf{d}}}$. Next, fix any $i \in \{1,\dots,N\}$
and let $\widehat{v}_{i}$ be the vector obtained at the $i^{th}$ site in
Step~15 of Algorithm~\ref{algo:Distributed_KSVD} during the $(t_p + 1)$
iteration of distributed power method. Notice that $\widehat{v}_{i}$ can be
expressed as $\widehat{v}_{i} = \widehat{v} + \epsilon_{i,c}$, where
$\epsilon_{i,c}$ denotes the error introduced in $\widehat{v}$ at the
$i^{th}$ site due to finite number of consensus iterations. Next, define $r = \|v\|_2$ and
$\widehat{r}_{i} = \|\widehat{v}_{i}\|_2$ and notice that
$q_{\textsf{c}}^\prime-q_{i,\textsf{d}}^\prime
=v(r^{-1}-\widehat{r}^{-1}_{i})+(v-\widehat{v}_{i})\widehat{r}^{-1}_{i}$. It
therefore follows from the triangle inequality that
\begin{equation}
\label{eq:BasicInequality}
\begin{aligned}
\|q_{\textsf{c}}^\prime-q_{i,\textsf{d}}^\prime\|_2 \leq \|v\|_2|r^{-1}-\widehat{r}^{-1}_{i}|+\|v-\widehat{v}_{i}\|_2\widehat{r}^{-1}_{i}.
\end{aligned}
\end{equation}

We now need to bound $\|v\|_2$, $|r^{-1}-\widehat{r}^{-1}_{i}|$,
$\|v-\widehat{v}_{i}\|_2$, and $\widehat{r}^{-1}_{i}$. To this end, notice
that $v-\widehat{v}_{i}
=\left[\sum_{i=1}^{N}{M_{i}(q_{\textsf{c}}-q_{i,\textsf{d}})}\right]-\epsilon_{i,c}$.
It also follows from Proposition~\ref{theorem:ConsensusError} and some
manipulations that $\|\epsilon_{i,c}\|_2 \leq \delta \gamma_p \sqrt{N}$. We
therefore obtain
\begin{equation}
\label{eq:boundV}
\begin{aligned}
\|v-\widehat{v}_{i}\|_2 \leq \sum_{i=1}^{N}{\|M_{i}\|_2 \|q_{\textsf{c}}-q_{i,\textsf{d}}\|_2}+\ConsensusError\gamma \sqrt{N}.
\end{aligned}
\end{equation}
Next, notice $|r^{-1}-\widehat{r}_{i}^{-1}| = |r -
\widehat{r}_{i}|r^{-1}\widehat{r}_{i}^{-1}$ and further it can be shown that
$|r-\widehat{r}_{i}| \leq r^{-1}|\widehat{r}_{i}^{2}-r^{2}|$. Now,
{
$|\widehat{r}_{i}^{2}-r^{2}|=|\widehat{v}_{i}^{\tT}\widehat{v}_{i}-v^{\tT}v| \leq \|\widehat{v}_{i}-v\|_2 (\|\widehat{v}_{i}\|_2 + \|v\|_2).$}
Since $\widehat{v}_{i} = \widehat{v} + \epsilon_{i,c}$, it can also be shown
that $\|\widehat{v}_{i}\|_2 \leq \alpha_p + \delta \gamma_p \sqrt{N}$. In
addition, we have $\|v\|_2 \leq \alpha_p$. Combining these facts with
\eqref{eq:boundV}, we get {\small\begin{align} \label{eq:rSquaredBound2}
&|\widehat{r}_{i}^{2}-r^{2}| \nonumber\\
&\leq (2\alpha_p +\delta \gamma_p \sqrt{N})\left(\sum_{i=1}^{N}{\|M_{i}\|_2 \|q_{\textsf{c}}-q_{i,\textsf{d}}\|_2}+\delta\gamma_p\sqrt{N}\right),\nonumber\\
    &\leq (2\alpha_p +\delta \gamma_p \sqrt{N})\left(\alpha_p \max_i \|q_{\textsf{c}}-q_{i,\textsf{d}}\|_2+\delta\gamma_p \sqrt{N}\right).
\end{align}}
We can now use this inequality to obtain $|r^{-1}-\widehat{r}_{i}^{-1}| \leq
\widehat{r}_{i}^{-1}\beta_p^{2}(2\alpha_p +\delta \gamma_p \sqrt{N})(\alpha_p
\max_{i}\|q_{\textsf{c}}-q_{i,\textsf{d}}\|_2+\delta\gamma_p \sqrt{N})$.

The only remaining quantity we need to bound is $\widehat{r}_{i}^{-1}$. To
this end, notice that $|r - \widehat{r}_{i}| \geq
(r^{-1})^{-1}-(\widehat{r}_{i}^{-1})^{-1}$. Since $|r-\widehat{r}_{i}| \leq
r^{-1}|\widehat{r}_{i}^{2}-r^{2}|$, we obtain from \eqref{eq:rSquaredBound2}
that
{\begin{align*}
(r^{-1})^{-1}-(\widehat{r}_{i}^{-1})^{-1} \leq &\alpha_p r^{-1}(2\alpha_p+\delta \gamma_p \sqrt{N})\\
&\left(\max_{i}\|q_{\textsf{c}}-q_{i,\textsf{d}}\|_2+\frac{\delta \gamma_p \sqrt{N}}{\alpha_p}\right).
\end{align*}}
It then follows from the lemma's assumptions along with some algebraic
manipulations that $\widehat{r}_{i}^{-1} \leq 2\beta_p$. Finally, plugging
the bounds on $\widehat{r}_{i}^{-1}$, $|r^{-1} - \widehat{r}_{i}^{-1}|$,
$\|v\|_2$, and $\|v-\widehat{v}_{i}\|_2$ in \eqref{eq:BasicInequality}, we
obtain
\begin{align*}
&\|q_{\textsf{c}}^\prime-q_{i,\textsf{d}}^\prime\|_2 \\
&\leq 2\alpha_p \beta_p^{3}\left(\alpha_p \max_{i}\|q_{\textsf{c}}-q_{i,\textsf{d}}\|_2+\delta \gamma_p \sqrt{N}\right)\\
&(2\alpha_p +\delta \gamma_p \sqrt{N})+ 2\beta_p\left(\alpha_p \max_{i}\|q_{\textsf{c}}-q_{i,\textsf{d}}\|_2+\delta \gamma_p \sqrt{N}\right)\\
&=\left(4\alpha_p^3 \beta_p^{3} + 2\alpha_p^{3} \beta_p^{3} \frac{\delta \gamma_p \sqrt{N}}{\alpha_p}+ 2\alpha_p \beta_p\right)\\
&\quad\left(\max_{i}\|q_{\textsf{c}}-q_{i,\textsf{d}}\|_2+\frac{\delta \gamma_p \sqrt{N}}{\alpha_p}\right).
\end{align*}
Finally, $\frac{\delta \gamma_p \sqrt{N}}{\alpha_p} \leq
\big(\frac{\epsilon}{2}\big)^{3T_p} < 1$ since ($i$) $\delta =
\frac{\alpha_p}{ \gamma_p
\sqrt{N}}\big(\frac{\epsilon}{2\alpha_p\beta_p}\big)^{3T_p}$, ($ii$)
$\epsilon < 1$, and ($iii$) $\alpha_p r^{-1} \geq 1$, which implies $\alpha_p
\beta_p \geq 1$. Plugging this into the above expression and noting that
$\alpha_p \beta_p \leq \alpha_p^3 \beta_p^{3}$, we obtain the claimed result.
\end{IEEEproof}

Lemma~\ref{le:ErrorInIteration} provides an understanding of the error
accumulation in the distributed power method due to finite number of
consensus iterations in each power method iteration. And while the factor of
$(2\alpha_p \beta_p)^{3}$ in the lemma statement might seem discouraging, the
fact that the distributed power method starts with a zero error helps keep
the total error in control. We now formally argue this in the proof of
Theorem~\ref{theorem:DistributedPowerMethod} below.

\begin{IEEEproof}[Proof of Theorem~\ref{theorem:DistributedPowerMethod}]
We begin by defining $q_{\textsf{c}}$ as the estimate of $u_1$ obtained using
$T_p$ iterations of the centralized power method that is initialized with the
same $q^{init}$ as the distributed power method. Next, fix an $i \in
\{1,\dots,N\}$ and notice that
\begin{align}
\label{eqn:dpm_triIneq}
\left\|u_1 u_1^\tT - \widehat{q}_i\widehat{q}_i^\tT\right\|_2 \leq
    \|u_1u_1^{\tT}-q_{\textsf{c}}q_{\textsf{c}}^{\tT}\|_2+\|q_{\textsf{c}}q_{\textsf{c}}^{\tT}-\widehat{q}_i {\widehat{q}_i}^{\tT}\|_2.
\end{align}
The convergence rate of the centralized power method is well studied and can
be expressed as \cite{golub2012matrix}
\begin{equation}
\|u_1 u_1^{\tT}-q_{\textsf{c}}q_{\textsf{c}}^{\tT}\|_2 \leq \tan{(\theta)}
\left|\frac{\lambda_{2}}{\lambda_{1}}\right|^{T_p}.
\end{equation}
In order to bound $\|q_{\textsf{c}}q_{\textsf{c}}^{\tT}-\widehat{q}_i
{\widehat{q}_i}^{\tT}\|_2$, we make use of Lemma~\ref{le:ErrorInIteration}.
To invoke this lemma, we first need to show that the main assumption of the
lemma holds for all iterations $t_p \leq (T_p - 1)$. We start with $t_p = 0$
for this purpose and note that $q_{\textsf{c}}^{(0)} = {\widehat{q}_i}^{(0)}
= q^{init}$, which trivially implies
{$\|q_{\textsf{c}}^{(0)}-{\widehat{q}_i}^{(0)}\|_2 +
\frac{\delta\gamma_p \sqrt{N}}{\alpha_p} \leq (\tfrac{\epsilon}{2})^{3T_p},$}
where $\delta$ is as defined in Lemma~\ref{le:ErrorInIteration}. Further,
under the assumptions of the theorem, it can be shown through elementary
algebra that $\big(\frac{\epsilon}{2}\big)^{3T_p} \leq
\frac{1}{2\alpha_p\beta_p^2(2\alpha_p+\delta\gamma_p \sqrt{N})}$. We now
invoke mathematical induction and claim that the main assumption of
Lemma~\ref{le:ErrorInIteration} is satisfied for all $t_p \leq m < T_p$. Then
we obtain from a recursive application of the statement of the lemma that for
$t_p = (m + 1)$, we have {\begin{align}\label{eqn:thm_pf_eqn1}
\hspace{-4em}&\|q_{\textsf{c}}^{(m+1)}-{\widehat{q}_i}^{(m+1)}\|_2 + \frac{\delta\gamma_p \sqrt{N}}{\alpha_p} \nonumber\\
&\quad\leq \frac{\delta \gamma_p \sqrt{N}}{\alpha_p} \sum_{i=0}^{m}{(2\alpha_p \beta_p)}^{3i}\stackrel{(a)}{\leq} 2\cdot\frac{\delta \gamma_p \sqrt{N}}{\alpha_p}(2\alpha_p \beta_p)^{3m}\nonumber\\
&\quad= 2\cdot\epsilon^{3T_p}\frac{(2\alpha_p \beta_p)^{3m}}{(2\alpha_p \beta_p)^{3T_p}}\stackrel{(b)}{\leq} \frac{1}{2\alpha_p\beta_p^{2}(2\alpha_p+\delta\gamma_p \sqrt{N})},
\end{align}}
where $(a)$ follows from the geometric sum and the fact that $(2\alpha_p
\beta_p)^{3} > 2$, while $(b)$ follows from the theorem assumptions and the
fact that $m < T_p$. We have now proved that the main assumption of
Lemma~\ref{le:ErrorInIteration} holds for all $t_p \leq (T_p-1)$. In order to
compute $\|q_{\textsf{c}}q_{\textsf{c}}^{\tT}-\widehat{q}_i
{\widehat{q}_i}^{\tT}\|_2$, therefore, we can recursively apply the result of
this lemma up to the $T_p^{th}$ iteration to obtain
\begin{align}
\|q_{\textsf{c}}-\widehat{q}_i\|_2 \leq \frac{\delta \gamma_p \sqrt{N}}{\alpha_p} \sum_{i=0}^{T_p}{(2\alpha_p \beta_p)^{3i}} \stackrel{(c)}{\leq} 2 \epsilon^{3T_p},
\end{align}
where $(c)$ follows from the same arguments as in \eqref{eqn:thm_pf_eqn1}.
The proof of the theorem now follows by noting the fact that
$\|q_{\textsf{c}}q_{\textsf{c}}^{\tT}-\widehat{q}_i {\widehat{q}_i}^{\tT}\|_2
\leq (\|q_{\textsf{c}}\|_2 + \|\widehat{q}_i\|_2)\|q_{\textsf{c}} -
\widehat{q}_i\|_2 \leq 4\epsilon^{3T_p}$.%
\end{IEEEproof}

\section{Proof of Theorem~\ref{thm:AllIteration}}\label{app:proof_thm_Bik}
Notice from Algorithm~\ref{algo:Distributed_KSVD} that sparse coding is always performed before update of the first dictionary atom. However, we do not perform sparse coding before updating any other dictionary atom. Due to this distinction, we answer how error is accumulated in matrix $E_{i,k,R}^{(t)}$ for first dictionary atom differently than for any other dictionary atom. In the following, we first provide an overview of how to bound $\|B_{i,k+1,R}^{(t)}\|_2$ when we know a bound on $\|B_{i,k,R}^{(t)}\|_2$. Then we will talk about bounding $\|B_{i,1,R}^{(t+1)}\|_2$ when we know bounds on $\{\|B_{i,j,R}^{(t)}\|_2\}_{j=1}^{K}$.

Recall from Step.~5 in Algorithm~\ref{algo:Distributed_KSVD} that
$\widehat{E}_{i,k,R}^{(t)}=
Y_{i}\widetilde{\Omega}_{i,k}^{(t)}~-\sum_{j=1}^{k-1}{\widehat{d}_{i,j}^{(t)}
\widehat{x}_{i,j,T}^{(t)}\widetilde{\Omega}_{i,k}^{(t)}}-\sum_{j=k+1}^{K}{\widehat{d}_{i,j}^{(t-1)}\widetilde{x}_{i,j,T}^{(t)}\widetilde{\Omega}_{i,k}^{(t)}}$. Now, if one assumes that $\widetilde{\Omega}_k^{(t)}=\Omega_k^{(t)}$, which we will argue is true, then the error in $E_{i,k,R}^{(t)}$ is due to errors in $\{x_{i,j,T,R}^{(t)}\}_{j=1}^{K}$ and $\{d_{j}^{(t)}\}_{j=1}^{K}$. Infact, we will show that $\|B_{i,k+1,R}^{(t)}\|_2$ can be bounded by knowing bounds on errors in $\widehat{d}_{i,k}^{(t)}$ and $x_{i,k,T,R}^{(t)}$ only. Next, recall from Step.~19 in Algorithm~\ref{algo:Distributed_KSVD} that $\widehat{x}_{i,k,R}^{(t)}=\widehat{d}_{i,k}^{(t)}\widehat{E}_{i,k,R}^{(t)^{\tT}}$, which means we only need to know a bound on $d_{k}^{(t)}$ to bound $\|B_{i,k+1,R}^{(t)}\|_2$. Another challenge for us will be to bound error in $d_{k}^{(t)}$ from a given bound on $\|B_{i,k,R}^{(t)}\|_2$. We will accomplish this by noting that there are two sources of error in $\widehat{d}_{k}^{(t)}$. The first source is the difference in eigenvectors of $\widehat{E}_{k,R}^{(t)} \widehat{E}_{k,R}^{(t)^{\tT}}$ and  $E_{k,R}^{(t)} E_{k,R}^{(t)^{\tT}}$. We will bound this difference using Proposition~\ref{prop:PerturbedEigenvector} in Appendix~\ref{app:prior_results}. In order to use this proposition, we will need a bound on  $\|\widehat{E}_{k,R}^{(t)}\widehat{E}_{k,R}^{(t)^{\tT}}- E_{k,R}^{(t)}E_{k,R}^{(t)^{\tT}}\|_F$, which we will also prove using a given bound on $\|B_{i,k,R}^{(t)}\|_2$ (Lemma~\ref{le:PerturbedEigenvectorAssumption}). The second source of error in $\widehat{d}_{k}^{(t)}$ is the error in eigenvector computation, which in our case is due to the distributed power method. It follows from Theorem~\ref{theorem:DistributedPowerMethod} and statement of Theorem~\ref{thm:AllIteration} that this error is bounded by $\varepsilon$. Combining these two sources of error, we will first bound the error in $\widehat{d}_k^{(t)}$ (Lemma~\ref{le:B_to_d}), {and then using this we will finally bound $\|B_{i,k+1,R}^{(t)}\|_2$ (Lemma~\ref{le:Bk_To_Bk1}).}

In order to bound $\|B_{i,1,R}^{(t+1)}\|_2$ when we know bounds on $\{\|B_{i,j,R}^{(t)}\|_2\}_{j=1}^{K}$, the difference from previous case is that now we can not write sparse code $\{\widehat{x}_{i,j,T}^{(t+1)}\}_{j=1}^{K}$ in terms of dictionary atoms $\{\widehat{d}_{i,j}^{(t)}\}_{j=1}^{K}$. Therefore, in addition to bounding errors in dictionary atoms $\{\widehat{d}_{i,j}^{(t)}\}_{j=1}^{K}$, we also need to bound errors in sparse codes due to perturbations in dictionaries after iteration $t$. Since we know $\{\|B_{i,k,R}^{(t)}\|_2\}_{j=1}^{K}$, we can use the bounds on $\{\widehat{d}_{i,j}^{(t)}\}_{j=1}^{K}$ derived earlier (Lemma~\ref{le:B_to_d}). Next, using error bounds on $\{\widehat{d}_{i,j}^{(t)}\}_{j=1}^{K}$, we can use Proposition~\ref{prop:SparseCodingBound} in Appendix~\ref{app:prior_results} to bound errors in $\{\widehat{x}_{i,j,T}^{(t+1)}\}_{j=1}^{K}$. Finally, using these error bounds on $\{\widehat{d}_{i,j}^{(t)}\}_{j=1}^{K}$ and $\{\widehat{x}_{i,j,T}^{(t+1)}\}_{j=1}^{K}$ we will bound $\|B_{i,1,R}^{(t+1)}\|_2$ (Lemma~\ref{le:BK_To_B1}). This will be followed by the remaining proof of Theorem~\ref{thm:AllIteration}.

Our first result in support of Theorem~\ref{thm:AllIteration} shows that the assumption of Proposition~\ref{prop:PerturbedEigenvector} in Appendix~\ref{app:prior_results} is satisfied under certain conditions, which will make it possible for us to bound the difference in the principal eigenvector of $E_{k,R}^{(t)} E_{k,R}^{(t)^{\tT}}$ and $\widehat{E}_{k,R}^{(t)} \widehat{E}_{k,R}^{(t)^{\tT}}$.
\begin{lemma}\label{le:PerturbedEigenvectorAssumption}
Let $\Omega_{i,k}^{(t)}$, $\widetilde{\Omega}_{i,k}^{(t)}$, $\varepsilon$ and
$\zeta$ be as defined in Theorem~\ref{thm:AllIteration}. Fix $\delta_d$ as in
Theorem~\ref{thm:main_result}, and suppose (i) P1--P3 are satisfied, (ii)
$\Omega_{i,k}^{(t)}=\widetilde{\Omega}_{i,k}^{(t)}$, and (iii) $\varepsilon
\leq \frac{\delta_d}{8N\sqrt{n}C_3(1+\zeta)^{T_d-1} C_4^2 (8C_3 N C_4^2 +
5)^{2(T_d K-2)}}$. Then $\forall i\in\{1,\dots,N\}$ and for any
{\small$t\in\{1,\cdots,T_d\}$ and $k\in\{1,\cdots,K\}$, if}
\begin{align*}
\|B_{i,k,R}^{(t)}\|_2 \leq
 \begin{cases}
	  0 ,\qquad t=1, k=1,\\
    \varepsilon (1+\zeta)^{t-1} C_4 (8C_3 N C_4^2+5)^{(t-1)K+k-2},\; \text{o.w},
	\end{cases}
\end{align*}
then $\Delta M_k^{(t)}=E_{k,R}^{(t)}E_{k,R}^{(t)^{\tT}}-\widehat{E}_{k,R}^{(t)}\widehat{E}_{k,R}^{(t)^{\tT}}$ is bounded as $\|\Delta M_k^{(t)}\|_F \leq \frac{1}{5C_3}.$
\end{lemma}
\begin{IEEEproof}
Since our starting dictionaries are same, therefore, for $(t,k)=(1,1)$ we have $E_{1,R}^{(1)}=\widehat{E}_{1,R}^{(1)}$, which means $\Delta M_k = 0$. Hence, claim is true for $(t, k)=(1,1)$. In the following, proof is provided for the claim for case $(t,k)\neq 1$.

{Substituting $B_{i,k,R}^{(t)}$ in the definition of $\Delta
M_{k}^{(t)}$, we get}
\begin{align*}
\Delta M_k^{(t)} = \sum_{i=1}^{N}{E_{i,k,R}^{(t)}B_{i,k,R}^{{(t)}^{\tT}}+B_{i,k,R}^{(t)}E_{i,k,R}^{{(t)}^{\tT}}+B_{i,k,R}^{(t)}B_{i,k,R}^{{(t)}^{\tT}}}.
\end{align*}
Simple algebraic manipulations, along with submultiplicativity of matrix 2-norm, result in
\begin{align}
\|\Delta M_k^{(t)}\|_2 &\leq 2\sum_{i=1}^{N}{\left(\|E_{i,k,R}^{(t)}\|_2 \|B_{i,k,R}^{(t)}\|_2+\|B_{i,k,R}^{(t)}\|_2^2\right)}\nonumber\\
&\leq 2N\max_{i}{\left(C_4 \|B_{i,k,R}^{(t)}\|_2+\|B_{i,k,R}^{(t)}\|_2^2\right)},\label{eq:NormDeltaE}
\end{align}
where the last inequality is due to~\eqref{eqn:C_4_defn}. Now, using the assumptions on bound of $\|B_{i,k,R}^{(t)}\|_2$ and $\varepsilon$, {we get}
\begin{align*}
&\|\Delta M_k^{(t)}\|_2\\
&\leq 2N \varepsilon (1+\zeta)^{t-1}\left(C_4^2 (8 C_3 N C_4^2 + 5)^{(t-1)K+k-2}\right.\\
 &\quad\left.+\varepsilon (1+\zeta)^{t-1} C_4^2 (8 C_3 N C_4^2 + 5)^{2(t-1)K+2k-4}\right)\\
&\leq 2N \varepsilon (1+\zeta)^{t-1}\left(C_4^2 (8 C_3 N C_4^2 + 5)^{(t-1)K+k-2} \right.&\\
&\left.\quad+ \frac{1}{8N\sqrt{n}C_3}\frac{(1+\zeta)^{t-1} C_4^2 (8 C_3 N C_4^2 + 5)^{2(t-1)K+2k-4}\delta_d}{(1+\zeta)^{T_d-1} C_4^2 (8C_3 N C_4^2 + 5)^{2(T_d K-2)K}}\right)\\
&\stackrel{(a)}\leq 2N \varepsilon (1+\zeta)^{t-1}\left(C_4^2 (8 C_3 N C_4^2 + 5)^{(t-1)K+k-2}+ \frac{1}{8N\sqrt{n}C_3}\right)\\
&\leq 4N \varepsilon (1+\zeta)^{t-1}C_4^2 (8 C_3 N C_4^2 + 5)^{(t-1)K+k-2},
\end{align*}
where (a) is true because $\frac{(1+\zeta)^{t-1} C_4^2 (8 C_3 N C_4^2 + 5)^{2(t-1)K+2k-4}\delta_d}{(1+\zeta)^{T_d-1} C_4^2 (8C_3 N C_4^2 + 5)^{2(T_d K-2)}}\leq 1$. Finally, using once again the assumption on $\varepsilon$, performing algebraic manipulations and using the fact that $\delta_d \leq 1$ , we get
{\begin{align*}
\|\Delta M_k^{(t)}\|_2 &\leq \frac{(8 C_3 N C_4^2 + 5)^{(t-1)K+k-2}}{2\sqrt{n}C_3(8 C_3 N C_4^2 + 5)^{2(T_d K-2)}}\\
&\leq \frac{1}{\sqrt{n}(8 C_3 N C_4^2 + 5)^{(T_d K-2)}}\leq \frac{1}{\sqrt{n}(5 C_3)}.
\end{align*}}
{\small Now using the fact that $\text{rank}(\Delta M_k^{(t)})\leq n$, we get
$\|\Delta M_k^{(t)}\|_F \leq \sqrt{\text{rank}(\Delta M_k^{(t)})}\|\Delta M_k^{(t)}\|_2\leq\sqrt{n}\|\Delta M_k^{(t)}\|_2\leq \frac{1}{5 C_3}$.}
\end{IEEEproof}

{We are now ready to prove that if we know a bound on $\|B_{i,k,R}^{(t)}\|_2$ then we can bound the error in dictionary atom $\widehat{d}_{i,k}^{(t)}$. This result is given in the following lemma.}

\begin{lemma}\label{le:B_to_d}
Let $\Omega_{i,k}^{(t)}$, $\widetilde{\Omega}_{i,k}^{(t)}$, $\varepsilon$ and $\zeta$ be as defined in Theorem~\ref{thm:AllIteration}, also perform $T_c$ consensus iterations as given in Theorem~\ref{thm:AllIteration}. Now fix $\delta_d$ as in Theorem~\ref{thm:main_result}, and suppose (i) P1--P3 are satisfied, (ii) $\Omega_{i,k}^{(t)}=\widetilde{\Omega}_{i,k}^{(t)}$, and (iii) $\varepsilon \leq \frac{\delta_d}{8N\sqrt{n}C_3(1+\zeta)^{T_d-1} C_4^2 (8C_3 N C_4^2 + 5)^{2(T_d K-2)}}$. Then for all $i\in \{1,\dots,N\}$ and for any $t\in\{1,2,\cdots,T_d\}$ and $k\in\{1,2,\cdots,K\}$ if we know
\begin{align*}
\|B_{i,k,R}^{(t)}\|_2 \leq
 \begin{cases}
	  0 ,\qquad t=1, k=1,\\
    \varepsilon (1+\zeta)^{t-1} C_4 (8C_3 N C_4^2+5)^{(t-1)K+k-2},\; \text{o.w},
	\end{cases}
\end{align*}
{\footnotesize then, $\|\widehat{d}^{(t)}_{i,k}\widehat{d}^{(t)^{\tT}}_{i,k}-d^{(t)}_{k}d^{(t)^{\tT}}_{k}\|_2 \leq \varepsilon (1+\zeta)^{t-1} (8C_3 N C_4^2 + 5)^{(t-1)K+k-1}$.}
\end{lemma}
\begin{IEEEproof}
To prove this lemma we first need to decompose error in dictionary atom into two different components i.e., error in principal eigenvector due to perturbation in $E_{k,R}^{(t)} E_{k,R}^{(t)^{\tT}}$ and error due to distributed power method. Let $d_{k}^{(t)}$ be the updated $k^{th}$ atom of centralized dictionary at iteration $t$, which is the principal eigenvector of $E_{k,R}^{(t)}E_{k,R}^{(t)^{\tT}}$. In cloud $K$-SVD, $\widehat{d}_{i,k}^{(t)}$ corresponds to the principal eigenvector estimate of $\widehat{E}_{k,R}^{(t)}\widehat{E}_{k,R}^{(t)^{\tT}}$ obtained at the $i^{th}$ site. Let us denote the true principal eigenvector of $\widehat{E}_{k,R}^{(t)}\widehat{E}_{k,R}^{(t)^{\tT}}$ by $\widetilde{d}_{k}^{(t)}$ and let $\widehat{d}_{i,k}^{(t)}$ be the eigenvector of $\widehat{E}_{k,R}^{(t)}\widehat{E}_{k,R}^{(t)^{\tT}}$ computed using distributed power method at the $i^{th}$ site. Using this notation, notice that
{$\|d_k^{(t)}d_k^{(t)^{\tT}}-\widehat{d}_{i,k}^{(t)}\widehat{d}_{i,k}^{(t)^{\tT}}\|_2 \leq \|d_k^{(t)}d_k^{(t)^{\tT}}-\widetilde{d}_{k}^{(t)}\widetilde{d}_{k}^{(t)^{\tT}}\|_2
+ \|\widetilde{d}_{k}^{(t)}\widetilde{d}_{k}^{(t)^{\tT}}-\widehat{d}_{i,k}^{(t)}\widehat{d}_{i,k}^{(t)^{\tT}} \|_2$,}
where the first term is due to perturbation in $E_{k,R}^{(t)}E_{k,R}^{(t)^{\tT}}$ and the second term is due to imperfect power method and consensus iterations. We can now use Theorem~\ref{theorem:DistributedPowerMethod} to obtain
{\begin{align*}
&\|d_k^{(t)}d_k^{(t)^{\tT}}-\widehat{d}_{i,k}^{(t)}\widehat{d}_{i,k}^{(t)^{\tT}}\|_2 \\
&\leq \|d_k^{(t)}d_k^{(t)^{\tT}}-\widetilde{d}_{k}^{(t)}\widetilde{d}_{k}^{(t)^{\tT}}\|_2+ \tan{(\widehat{\theta}_k^{(t)})}\left(\frac{\widehat{\lambda}_{2,k}^{(t)}}{\widehat{\lambda}_{1,k}^{(t)}}\right)^{T_p}+4\epsilon^{3T_p}\nonumber\\
&\stackrel{(a)}\leq \|d_k^{(t)}d_k^{(t)^{\tT}}-\widetilde{d}_{k}^{(t)}\widetilde{d}_{k}^{(t)^{\tT}}\|_2+ \mu \nu^{T_p}+4\epsilon^{3T_p}\\
&\stackrel{(b)}= \|d_k^{(t)}d_k^{(t)^{\tT}}-\widetilde{d}_{k}^{(t)}\widetilde{d}_{k}^{(t)^{\tT}}\|_2+ \varepsilon,
\end{align*}}
where (a) is due to definition of parameters $\mu$ and $\nu$ in Theorem~\ref{thm:main_result}, and (b) {is due to} definition of $\varepsilon$ in Theorem~\ref{thm:AllIteration}.

Next, for symmetric matrices $M_k^{(t)}=\sum_i{E_{i,k,R}^{(t)}
E_{i,k,R}^{(t)^{\tT}}}$ and
$\widehat{M}_k^{(t)}=\sum_i{\widehat{E}_{i,k,R}^{(t)}
\widehat{E}_{i,k,R}^{(t)^{\tT}}}$ such that
$\widehat{M}_k^{(t)}=M_k^{(t)}+\Delta M_k^{(t)}$, we can use
Lemma~\ref{le:PerturbedEigenvectorAssumption} and
Proposition~\ref{prop:PerturbedEigenvector} to find a bound on deviation in
principal eigenvector of $M_k^{(t)}$ due to perturbation $\Delta M_k^{(t)}$.
Since we have from Lemma~\ref{le:PerturbedEigenvectorAssumption} that
$\|\Delta M_k^{(t)}\|_F\leq \frac{1}{5 C_3}$, it follows from
Proposition~\ref{prop:PerturbedEigenvector} that
{\begin{align}
\label{eq:dkdkT}
&\hspace{-2em}\|d_k^{(t)}d_k^{(t)^{\tT}}-\widehat{d}_{i,k}^{(t)}\widehat{d}_{i,k}^{(t)^{\tT}}\|_2 \leq 4C_3 \|\Delta M_{k}^{(t)}\|_2+\varepsilon \nonumber\\
&\leq 8C_3 N\max_{i}{\left(C_4
\|B_{i,k,R}^{(t)}\|_2+\|B_{i,k,R}^{(t)}\|_2^2\right)}+\varepsilon,
\end{align}}
where the last inequality is due to~\eqref{eq:NormDeltaE}. Now using the bound on $\|B_{i,k,R}^{(t)}\|_2$ in the lemma statement, it can be shown using some algebraic manipulations that
{\begin{align*}
&\|\widehat{d}^{(t)}_{i,k}\widehat{d}^{(t)^{\tT}}_{i,k}-d^{(t)}_{k}d^{(t)^{\tT}}_{k}\|_2\\ &\quad\leq \varepsilon (1+\zeta)^{t-1} C_4\left(8 C_3 N C_4^2 (8C_3 N C_4^2 + 5)^{(t-1)K+k-2} \right.\\
&\quad\left.+ 8\varepsilon (1+\zeta)^{t-1} C_4 C_3 N (8C_3 N C_4^2 +5)^{2(t-1)K+2k-4}+1 \right).
\end{align*}}
The claim in the lemma now follows by replacing the bound on $\varepsilon$ in the parentheses of the above inequality, followed by some manipulations.
\end{IEEEproof}

The next lemma shows that if we know bounds on errors in $\{\widehat{E}_{i,k,R}^{(t)}\}_{k=1}^{K}$ for any $t$ then we can bound the error in $\widehat{E}_{i,1,R}^{(t+1)}$.
\begin{lemma}\label{le:BK_To_B1}
Let $\Omega_{i,k}^{(t)}$, $\widetilde{\Omega}_{i,k}^{(t)}$, $\varepsilon$ and
$\zeta$ be as defined in Theorem~\ref{thm:AllIteration}, also perform $T_c$
consensus iterations as given in Theorem~\ref{thm:AllIteration}. Now fix
$\delta_d$ as in Theorem~\ref{thm:main_result} and suppose (i) P1--P3 are
satisfied, (ii) $\Omega_{i,k}^{(t+1)}=\widetilde{\Omega}_{i,k}^{(t+1)}$,
(iii)  $\Omega_{i,k}^{(t)}=\widetilde{\Omega}_{i,k}^{(t)}$ , and (iv)
$\varepsilon \leq \frac{\delta_d}{8N\sqrt{n}C_3(1+\zeta)^{T_d-1} C_4^2 (8C_3
N C_4^2 + 5)^{2(T_d K-2)}},$ then for any $t\in\{1,\cdots,T_d-1\}$, and for
all $k\in\{1,\cdots,K\}$ and $i \in \{1,\cdots,N\}$, if
$\|B_{i,k,R}^{(t)}\|_2\leq \varepsilon (1+\zeta)^{t-1}C_4 (8C_3 C_4^2 N +
5)^{(t-1)K+k-2}$ then, {\small$\|B_{i,1,R}^{(t+1)}\|_2\leq \varepsilon (1+\zeta)^{t}
C_4 (8C_3 C_4^2 N + 5)^{tK-1}$}.
\end{lemma}
\begin{IEEEproof}
The error in $\widehat{E}_{i,1,R}^{(t+1)}$ is due to error in dictionary in the
previous iteration $t$ and sparse coding at the start of iteration $(t+1)$. Specifically,
$B_{i,1}^{(t+1)} = {E}_{i,1}^{(t+1)}-\widehat{E}_{i,1}^{(t+1)}=Y_i - \sum_{j=2}^{K}d_{j}^{(t)}x_{i,j,T}^{(t+1)} - Y_i + \sum_{j=2}^{K}\widehat{d}^{(t)}_{i,j}\widetilde{x}^{(t+1)}_{i,j,T}
$. It then follow that {$\|B_{i,1}^{(t+1)}\|_2 \leq \sum_{j=2}^{K}{\|\widehat{d}^{(t)}_{i,j}\widetilde{x}^{(t+1)}_{i,j,T}-d_{j}^{(t)}x_{i,j,T}^{(t+1)}\|_2}\leq \sum_{j=1}^{K}{\|\widehat{d}^{(t)}_{i,j}\widetilde{x}^{(t+1)}_{i,j,T}-d_{j}^{(t)}x_{i,j,T}^{(t+1)}\|_2}.$}
In reality we are interested in finding a bound on $\|B_{i,1,R}^{(t+1)}\|_2$. But since $\Omega_{i,k}^{(t+1)}=\widetilde{\Omega}_{i,k}^{(t+1)}$ we can define $B_{i,1,R}^{(t+1)}$ as
$B_{i,1,R}^{(t+1)}=\left(\sum_{j=2}^{K}{\left(Y_i-\widehat{d}^{(t)}_{i,j}\widetilde{x}^{(t+1)}_{i,j,T}\right)}-\sum_{j=2}^{K}{\left(Y_i -d_{j}^{(t)}x_{i,j,T}^{(t+1)}\right)}\right)\Omega_{i,1}^{(t+1)}$.
It can be seen from this definition that $B_{i,1,R}^{(t+1)}$ is a submatrix of $B_{i,1}^{(t+1)}$, which implies
{\small\begin{align}
\|B_{i,1,R}^{(t+1)}\|_2 \leq \|B_{i,1}^{(t+1)}\|_2\leq \sum_{j=1}^{K}{\|\widehat{d}^{(t)}_{i,j}\widetilde{x}^{(t+1)}_{i,j,T}-d_{j}^{(t)}x_{i,j,T}^{(t+1)}\|_2}.\label{eqn:B_1R}
\end{align}}
Now, defining $\widehat{d}^{(t)}_{i,j}= d_{j}^{(t)}+e_{i,j}^{(t)}$, where $e_{i,j}^{(t)}$ denotes the error in dictionary atom $d_{j}^{(t)}$, and substituting this in~\eqref{eqn:B_1R} we get
{\begin{align}
&\|B_{i,1,R}^{(t+1)}\|_2\nonumber\\
&\leq K \max_{j}\left({\|d_{j}^{(t)}\widetilde{x}^{(t+1)}_{i,j,T}-d_{j}^{(t)}x_{i,j,T}^{(t+1)}\|_2 + \|e_{i,j} \widetilde{x}^{(t+1)}_{i,j,T}\|_2}\right)\nonumber \\
&\leq K \max_{j}\left({\|\widetilde{x}^{(t+1)}_{i,j,T}-x^{(t+1)}_{i,j,T}}\|_2 + {\|\widehat{d}^{(t)}_{i,j}-d^{(t)}_{j}}\|_2 \|\widetilde{x}^{(t+1)}_{i,j,T}
\|_2\right)\nonumber \\
&= K \max_{j}\left({\|\widetilde{x}^{(t+1)}_{i,j,T}-x^{(t+1)}_{i,j,T}\|_2}\right.\nonumber\\
&\left.\quad+ \|\widehat{d}^{(t)}_{i,j}-d^{(t)}_{j}\|_2 \|\widetilde{x}^{(t+1)}_{i,j,T}+x^{(t+1)}_{i,j,T}-x^{(t+1)}_{i,j,T}\|_2\right)\nonumber\\
&\leq K \max_{j}\left({\|\widetilde{x}^{(t+1)}_{i,j,T}-x^{(t+1)}_{i,j,T}\|_2}(1 +
 \|\widehat{d}^{(t)}_{i,j}-d^{(t)}_{j}\|_2) \right.\nonumber\\
&\left.\quad+ \|\widehat{d}^{(t)}_{i,j}-d^{(t)}_{j}\|_2 \|x^{(t+1)}_{i,j,T}\|_2\right).\label{eqn:B_1Rb}
\end{align}}
Now, let {\small $X^{(t+1)}=\begin{bmatrix} X_{1}^{(t+1)} & X_{2}^{(t+1)} & \dots & X_{N}^{(t+1)} \end{bmatrix}\in \mathbb{R}^{K\times S}$} be the sparse coding matrix associated with the centralized $K$-SVD (see, e.g, Sec~\ref{subsec:reviewKSVD}). Notice that $x_{i,j,T}^{(t+1)}$ is the $j^{th}$ row of $X_{i}^{(t+1)}$. It then follows that
$$\|x_{i,j,T}^{(t+1)}\|_2\leq \sqrt{S_i} \|X_{i}^{(t+1)}\|_{\max} \leq \sqrt{S_i} \|X_{i}^{(t+1)}\|_1.$$
We therefore obtain under P1 that {$\|x_{i,j,T}^{(t+1)}\|_2\leq \sqrt{S_{\max}} \eta_{\tau,\max}$.}
Next, using the bound on $\|B_{i,k,R}^{(t)}\|_2$ and applying Lemma~\ref{le:B_to_d}, we get
$ \|\widehat{d}^{(t)}_{i,k}\widehat{d}^{(t)^{\tT}}_{i,k}-d^{(t)}_{k}d^{(t)^{\tT}}_{k} \|_2 \leq \varepsilon (1+\zeta)^{t-1} C_4(8C_3 N C_4^2 + 5)^{(t-1)K+k-1}.$
Now, under the assumption that both cloud $K$-SVD and centralized $K$-SVD use the same $d_{ref}$, we have $\widehat{d}_{i,k}^{(t)^{\tT}} d_{k}^{(t)}\geq 0$ and therefore it follows from Lemma~\ref{le:u_v} in Appendix~\ref{app:prior_results} that
\begin{align}
 \|\widehat{d}^{(t)}_{i,k}-d^{(t)}_{k} \|_2 &\leq \varepsilon \sqrt{2}(1+\zeta)^{t-1} C_4(8C_3 N C_4^2 + 5)^{(t-1)K+k-1}\nonumber\\
&\stackrel{(a)}\leq \sqrt{2}\delta_d\stackrel{(b)}\leq 1,\label{eqn:bound_d_2}
\end{align}
where (a) follows from the assumption on $\varepsilon$ and (b) is true for any fixed $\delta_d$ as defined in Theorem~\ref{thm:main_result}. Using this bound we can write
{\begin{align}\label{eqn:bound_D}
\|D^{(t)}-\widehat{D}^{(t)}_i\|_2 &\leq \|D^{(t)}-\widehat{D}^{(t)}_i\|_F=\sqrt{\sum_{j=1}^{K}{\|\widehat{d}^{(t)}_{i,j}-d^{(t)}_{j} \|_2^2}}\nonumber\\
&\hspace{-3em}\leq \sqrt{K}\max_{j\in\{1,\cdots,K\}}{\|\widehat{d}^{(t)}_{i,j}-d^{(t)}_{j} \|_2}\nonumber\\
&\hspace{-3em}\leq \sqrt{2K}{(1+\zeta)^{t-1} \varepsilon C_4(8C_3 N C_4^2 + 5)^{tK-1}}.
\end{align}}
{Furthermore, using lemma assumption on $\varepsilon$ we get
\begin{align}\label{eqn:bound_D_2}
\|D^{(t)}-\widehat{D}^{(t)}_i\|_2 \leq \sqrt{2K} \delta_d=\min\left\{\sqrt{K},\frac{C_1^2 \tau_{\min}}{44}\right\}.
\end{align}}
We can now use~\eqref{eqn:bound_D_2} and
Proposition~\ref{prop:SparseCodingBound} in Appendix~\ref{app:prior_results}
to bound $\|x_{i,j,T}^{(t+1)}-\widetilde{x}_{i,j,T}^{(t+1)}\|_2$
in~\eqref{eqn:B_1Rb}. Notice that Proposition~\ref{prop:SparseCodingBound}
assumes the error in dictionary to be smaller than $\frac{C_1^{2}
\tau_{\min}}{44}$, which is satisfied by~\eqref{eqn:bound_D_2}. Other
assumptions of Proposition~\ref{prop:SparseCodingBound} are satisfied due to
P1 and P2. Therefore, we get
{$\forall\;i\in\{1,\dots,N\}\text{ and }j\in\{1,\dots,S_i\},$
\begin{align}\label{eqn:bound_x_tilde}\|x_{i,j}^{(t+1)}-\widetilde{x}_{i,j}^{(t+1)}\|_2 \leq \frac{3 \sqrt{T_0}}{\tau_{\min} C_2}\|D^{(t)}-\widehat{D}^{(t)}_i\|_2.
\end{align}}
Now defining $X_{i}^{(t+1)}$ and $\widetilde{X}_i^{(t+1)}$ as before, we note that
{\begin{align}\label{eqn:x_ijT}
&\|x_{i,j,T}^{(t+1)}-\widetilde{x}_{i,j,T}^{(t+1)}\|_2\nonumber\\
&\leq \sqrt{S_{\max}}\|X_i^{(t+1)}-\widetilde{X}_i^{(t+1)}\|_{\max}\nonumber\\
&\leq \sqrt{S_{\max}}\max_{j\in\{1,\dots,S_i\}} {\|x_{i,j}^{(t+1)}-\widetilde{x}_{i,j}^{(t+1)}\|_2}\nonumber\\
&\leq \frac{3\sqrt{2K S_{\max}T_0} }{\tau_{\min} C_2} {\varepsilon (1+\zeta)^{t-1}} C_4(8C_3 N C_4^2 + 5)^{tK-1},
\end{align}}
{where the last inequality follows from~\eqref{eqn:bound_x_tilde}
and~\eqref{eqn:bound_D}.} Now using bounds on $\|x_{i,j,T}^{(t+1)}\|_2$
and~\eqref{eqn:x_ijT} we get the following from~\eqref{eqn:B_1Rb}:
{\begin{align*}
&\|B_{i,1,R}^{(t+1)}\|_2 \\
&\stackrel{(c)}\leq 2K \max_{j}{\|\widetilde{x}^{(t+1)}_{i,j,T}-x^{(t+1)}_{i,j,T}\|_2}+ \max_j \|\widehat{d}^{(t)}_{i,j}-d^{(t)}_{j}\|_2 \|x^{(t)}_{i,j,T}\|_2\\
&\stackrel{(d)}\leq 2K \frac{3\sqrt{S_{\max}T_0} }{\tau_{\min} C_2} \sqrt{2K}{\varepsilon (1+\zeta)^{t-1} C_4(8C_3 N C_4^2 + 5)^{tK-1}} \\
&\quad+ \varepsilon \sqrt{2}(1+\zeta)^{t-1} C_4(8C_3 N C_4^2 + 5)^{tK-1} \sqrt{S_{\max}}\eta_{\tau,\max}\\
&\stackrel{(e)}\leq  \varepsilon (1+\zeta)^{t} C_4(8C_3 N C_4^2 + 5)^{tK-1}.
\end{align*}}
Here, (c)--(d) follow by application of~\eqref{eqn:bound_d_2} and~\eqref{eqn:bound_D}, and (e) is by definition of $\zeta$.
\end{IEEEproof}
The last lemma that we need bounds $\|B_{i,k+1,R}^{(t)}\|_2$ when we have a bound on $\|B_{i,k,R}^{(t)}\|_2$.
\begin{lemma}\label{le:Bk_To_Bk1}
Let $\Omega_{i,k}^{(t)}$, $\widetilde{\Omega}_{i,k}^{(t)}$, $\varepsilon$ and
$\zeta$ be as defined in Theorem~\ref{thm:AllIteration}, also perform $T_c$
consensus iterations as given in Theorem~\ref{thm:AllIteration}. Now fix
$\delta_d$ as in Theorem~\ref{thm:main_result}, and suppose (i) P1--P3 are
satisfied, (ii) $\Omega_{i,k}^{(t)}=\widetilde{\Omega}_{i,k}^{(t)}$, and
(iii) $\varepsilon \leq \frac{\delta_d}{8N\sqrt{n}C_3
(1+\zeta)^{T_d-1}C_4^2(8C_3 N C_4^2 + 5)^{2(T_d K-2)}}$. For any fixed
$k\in\{1,\cdots,K\}$, $t\in\{1,\cdots,T_d\}$, and all $i \in \{1,\cdots,N\}$,
if $\|B_{i,k,R}^{(t)}\|_2\leq \varepsilon (1+\zeta)^{t-1}C_4 (8C_3 C_4^2 N +
5)^{(t-1)K+k-2}$
 then $\|B_{i,k+1,R}^{(t)}\|_2\leq \varepsilon (1+\zeta)^{t-1}C_4 (8C_3 C_4^2 N + 5)^{(t-1)K+k-1}$.
\end{lemma}
\begin{IEEEproof}
Recall once again that we can write
{\begin{align*}
&\hspace{-3.5em}B_{i,k+1,R}^{(t)}=\widehat{E}_{i,k+1,R}^{(t)}-E_{i,k+1,R}^{(t)} \nonumber \\
&= \sum_{j=k+2}^{K}\left({d_{j}^{(t-1)} x_{i,j,R}^{(t)} - \widehat{d}_{i,j}^{(t-1)} \widetilde{x}_{i, j,R}^{(t)}}\right)\nonumber\\
&\quad - \sum_{j=1}^{k}{\left(\widehat{d}_{i,j}^{(t)}\widehat{x}_{i,j,R}^{(t)} - d_j^{(t)} x_{i,j,R}^{(t)}\right)}, \nonumber
\end{align*}
now using relation $\widehat{x}_{i,k,R}^{(t)}=\widehat{d}_{i,k}^{(t)^{\tT}}\widehat{E}_{i,k,R}^{(t)}$ and doing some rearrangements we get,
\begin{align*}
B_{i,k+1,R}^{(t)}&= \widehat{d}_{i,k}^{(t)}\widehat{d}_{i,k}^{(t)^{\tT}}\widehat{E}_{i,k,R}^{(t)}-d_{k}^{(t)} d_{k}^{(t)^{\tT}} E_{i,k,R}^{(t)}\nonumber\\
& \quad- \left({d_{k+1}^{(t-1)} x_{i,k+1,R}^{(t)} - \widehat{d}_{i,k+1}^{(t-1)} \widetilde{x}_{i, k+1,R}^{(t)}}\right) + B_{i,k,R}^{(t)}.
\end{align*}}
It then follows that
{\small \begin{align*}
&\|B_{i,k+1,R}^{(t)}\|_2\\
&\leq \|B_{i,k,R}^{(t)}\|_2 + \left\|\widehat{d}_{i,k}^{(t)}\widehat{d}_{i,k}^{(t)^{\tT}}(E_{i,k,R}^{(t)}+B_{i,k,R}^{(t)})-d_{k}^{(t)} d_{k}^{(t)^{\tT}} E_{i,k,R}^{(t)}\right\|_2 \\
&\quad+\left\|{d_{k+1}^{(t-1)} x_{i,k+1,R}^{(t)} - \widehat{d}_{i,k+1}^{(t-1)} \widetilde{x}_{i, k+1,R}^{(t)}}\right\|_2\\
&\stackrel{(a)}\leq 2\|B_{i,k,R}^{(t)}\|_2 + \|\widehat{d}_{i,k}^{(t)}\widehat{d}_{i,k}^{(t)^{\tT}} - d_{k}^{(t)} d_{k}^{(t)^{\tT}}\|_2 C_4 \\
&\quad+ \left\|{d_{k+1}^{(t-1)} x_{i,k+1,R}^{(t)} - \widehat{d}_{i,k+1}^{(t-1)} \widetilde{x}_{i, k+1,R}^{(t)}}\right\|_2\\
&\stackrel{(b)}\leq \varepsilon C_4 (1+\zeta)^{t-1} \left((8C_3 N C_4^2 + 5)^{(t-1)K+k-2}(8C_3 N C_4^2 + 3)\right. \\
&\left.\quad+ \varepsilon 8 C_3 N C_4^2 (1+\zeta)^{t-1} (8C_3 N C_4^2 + 5)^{2(t-1)K+2k-4} \right.\\
&\left.\quad+ \frac{1}{(1+\zeta)^{t-1}}\right). \\
\end{align*}}
Here (a) is due to the fact that $E_{i,k,R}^{(t)}$ is a submatrix of $E_{i,k}^{(t)}$ and the definition of $C_4$ in~\eqref{eqn:C_4_defn}, {(b) is obtained by applying~\eqref{eq:dkdkT}, using assumption on $\|B_{i,k,R}^{(t)}\|_2$ and finally using the same procedure as in Lemma~\ref{le:BK_To_B1} after~\eqref{eqn:B_1R} to bound $\sum_{j=1}^{K}\left\|{d_{j}^{(t-1)} x_{i,j, T,R}^{(t)} - \widetilde{d}_{i,j}^{(t-1)} \widetilde{x}_{i, j, T,R}^{(t)}}\right\|_2$.}
 The proof of the lemma now follows by using the assumption on $\varepsilon$ and some algebraic manipulations.
\end{IEEEproof}
The proof of Theorem~\ref{thm:AllIteration} now can be given by combining Lemmas~\ref{le:PerturbedEigenvectorAssumption}--~\ref{le:Bk_To_Bk1}. Since these lemmas require the supports of both centralized and distributed problems to be the same, the main challenge in proving Theorem~\ref{thm:AllIteration} lies in showing this fact.
\begin{IEEEproof}[Proof of Theorem~\ref{thm:AllIteration}]
We will prove this theorem by mathematical induction over $t$. To be specific, we will prove the following two cases:
\begin{enumerate}
\item For base case, we will show that the claim holds for $\|B_{i,k,R}^{(1)}\|_2\;\forall k\in\{1,2,\cdots,K\}$.
\item For induction step we assume that for any $q\in\{1,2,\cdots,T_d-1\}$ the claim is true for $\|B_{i,k,R}^{(q)}\|_2\;\forall k\in\{1,2,\cdots,K\}$ and $\Omega_{i,k}^{(q)}=\widetilde{\Omega}_{i,k}^{(q)}$. Then we need to show that $\Omega_{i,k}^{(q+1)}=\widetilde{\Omega}_{i,k}^{(q+1)}$ and claim holds for $\|B_{i,k,R}^{(q+1)}\|_2\;\forall k\in\{1,2,\cdots,K\}$.
\end{enumerate}

\noindent
\underline{\textbf{Base case: $t=1\;\forall\;k\in\{1,2,\cdots,K\}$ }}
To prove the base case, we will do mathematical induction over $k$ by fixing $t=1$. Hence, the first thing we need to prove is that the bound is true for $\|B_{i,1,R}^{(1)}\|_2$. Since both cloud $K$-SVD and Centralized $K$-SVD start with the same initial dictionary, we have $d_j^{(0)}=\widehat{d}_{i,j}^{(0)},\;\forall\;j\in\{1,2,\cdots,K\}$. Therefore, we get $ \Omega_{i,j}^{(1)}=\widetilde{\Omega}_{i,j}^{(1)},\;\forall\;j\in\{1,2,\cdots,K\}$.
It then follows that $B_{i,1,R}^{(1)} =E_{i,1,R}^{(1)}-\widehat{E}_{i,1,R}^{(1)}=\sum_{j=1}^{K-1}\left(d_j^{(0)}x_{i,j,T}^{(1)}\Omega_{i,j}^{(1)}-\widehat{d}_j^{(0)}\widetilde{x}_{i,j,T}^{(1)}\widetilde{\Omega}_{i,j}^{(1)}\right)=0$, thereby proving the claim.

Next, for induction argument we fix $k=p\in\{1,\dots,K-1\}$ for $t=1$. Then we need to show that it holds for $k=p+1$. {Using the induction assumption we have $\|B_{i,p,R}^{(1)}\|_2 \leq \varepsilon C_4 (8C_3 N C_4^2 + 5)^{p-2}$.} Since $\Omega_{i,j}^{(1)}=\widetilde{\Omega}_{i,j}^{(1)}$, we have $B_{i,p+1,R}^{(1)}=\widehat{E}_{i,p+1,R}^{(1)}-E_{i,p+1,R}^{(1)}$. This results in
{\small\begin{align}
\label{eq:Bp1_1}
&\|B_{i,p+1,R}^{(1)}\|_2\nonumber\\
&= \| \sum_{j=p+2}^{K}{\left(d_{j}^{(0)} x_{i,j,R}^{(1)} -\widehat{d}_{i,j}^{(0)} \widetilde{x}_{i, j,R}^{(1)}\right)}\nonumber\\
&\quad- \sum_{j=1}^{p}{\left(\widehat{d}_{i,j}^{(1)}\widehat{x}_{i,j,R}^{(1)}-d_{j}^{(1)} x_{i,j,R}^{(1)}\right)} \|_2 \nonumber\\
&\stackrel{(a)}= \| - \sum_{j=1}^{p}{\left(\widehat{d}_{i,j}^{(1)}\widehat{x}_{i,j,R}^{(1)} - d_j^{(1)} x_{i,j,R}^{(1)}\right)}\|_2 \nonumber \\
&= \| \widehat{d}_{i,p}^{(1)}\widehat{x}_{i,p,R}^{(1)} - d_p^{(1)} x_{i,p,R}^{(1)} + \sum_{j=1}^{p-1}{\left(\widehat{d}_{i,j}^{(1)}\widehat{x}_{i,j,R}^{(1)} - d_j^{(1)} x_{i,j,R}^{(1)}\right)}\|_2 \nonumber\\
&= \|\widehat{d}_{i,p}^{(1)}\widehat{x}_{i,p,R}^{(1)} - d_p^{(1)} x_{i,p,R}^{(1)} + B_{i,p,R}^{(1)}\|_2,
\end{align}}
where (a) is true because $d_{j}^{(0)} x_{i,j,R}^{(1)}-\widehat{d}_{i,j}^{(0)} \widetilde{x}_{i, j,R}^{(0)}=0$. Substituting $\widehat{x}_{i,p,R}^{(1)}=\widehat{d}_{i,p}^{(1)^{\tT}}\widehat{E}_{i,p,R}^{(1)}$, we get
{\small\begin{align*}
&\|B_{i,p+1,R}^{(1)}\|_2\\
&\leq 2\|B_{i,p,R}^{(1)}\|_2 + \|\widehat{d}_{i,p}^{(1)}\widehat{d}_{i,p}^{(1)^{\tT}} - d_{p}^{(1)} d_{p}^{(1)^{\tT}}\|_2 \|E_{i,p,R}^{(1)}\|_2\\
&\stackrel{(b)}\leq 2\|B_{i,p,R}^{(1)}\|_2 + \|\widehat{d}_{i,p}^{(1)}\widehat{d}_{i,p}^{(1)^{\tT}} - d_{p}^{(1)} d_{p}^{(1)^{\tT}}\|_2 C_4\\
&\stackrel{(c)}\leq 2\|B_{i,p,R}^{(1)}\|_2 + C_4\left(8C_3  N \max_{i}\left(\|B_{i,p,R}^{(1)}\|_2 C_4 + \|B_{i,p,R}^{(1)}\|_2^2 \right) + \varepsilon \right)\\
&\stackrel{(d)}\leq \varepsilon C_4 \left((8C_3 N C_4^2 + 5)^{p-2}(8C_3 N C_4^2 + 2)\right.\\
&\quad\left.+ \varepsilon 8 C_3 N C_4^2 (8C_3 N C_4^2 + 5)^{2p-4} + 1\right). \\
\end{align*}}
Here, (b) is true since $E_{i,k,R}^{(t)}$ is a submatrix of $E_{i,k}^{(t)}$ and due to the definition of $C_4$ in~\eqref{eqn:C_4_defn}, (c) is {due to}~\eqref{eq:dkdkT} and (d) follows from using the bound on $\|B_{i,p,R}^{(1)}\|_2$ and some manipulations. Now using  the assumption on $\varepsilon$, we get
{
$\|B_{i,p+1,R}^{(1)}\|_2\leq\varepsilon C_4 (8C_3 N C_4^2 + 5)^{p-1}.$
}

\noindent
{\textbf{Induction step: Bound on $\|B_{i,k,R}^{(q)}\|_2$ holds for}}
\underline{\textbf{$t=q\in\{1,\dots,T_d-1\}$ and $\forall\;k\in\{1,2,\cdots,K\}$.}}

We need to show the bound holds for $\|B_{i,k,R}^{(q+1)}\|_2\;\forall\;k\in\{1,\dots,K\}$.
To show this, we will be using induction argument over $k$ by fixing $t=q+1$. {As base case we bound $\|B_{i,1,R}^{(q+1)}\|_2$}. To bound $\|B_{i,1,R}^{(q+1)}\|_2$, we will be using Lemma~\ref{le:BK_To_B1}, which assumes $\widetilde{\Omega}_{i,1}^{(q+1)}=\Omega_{i,1}^{(q+1)}$. Using the induction assumptions, we get the following bound on error in dictionary $\widehat{D}_i^{(q)}$ using Lemma~\ref{le:B_to_d} and performing same steps as we carried out in Lemma~\ref{le:BK_To_B1} to get~\eqref{eqn:bound_d_2} and~\eqref{eqn:bound_D}: $\|D^{(q)}-\widehat{D}_{i}^{(q)}\|_2\leq \varepsilon \sqrt{2K} C_4 (8C_3 N C_4^2 + 5)^{(q-1)K+k-2}$.
Using the assumption on $\varepsilon$, we then have $\|D^{(q)}-\widehat{D}_{i}^{(q)}\|_2\leq \delta_d\sqrt{2K}$. It then follows from arguments similar to the ones made in Lemma~\ref{le:BK_To_B1} that $\Omega_{i,1}^{(q+1)}=\widetilde{\Omega}_{i,1}^{(q+1)}$. We can now use Lemma~\ref{le:BK_To_B1} to bound {$\|B_{i,1,R}^{(q+1)}\|_2 \leq \varepsilon (1+\zeta)^{q} C_4 (8N C_3 C_4^2 + 5 )^{qK-1}$.}
Having proved the base case, we now suppose that the claim is true for some $k=p\in\{1,\dots,K-1\}$. We then need to show it holds for $\|B_{i,p+1,R}^{(q+1)}\|_2$. That claim, however, simply follows from Lemma~\ref{le:Bk_To_Bk1}. This concludes the proof of theorem.
\end{IEEEproof}

\section{Proof of Theorem~\ref{thm:main_result}}\label{app:proof_mainthm}
To prove Theorem~\ref{thm:main_result} we need an upper bound on error in matrices $\widehat{E}_{i,k,R}^{(t)}$, which is given by Theorem~\ref{thm:AllIteration}. Applying Theorem~\ref{thm:AllIteration} to get a bound on the error in dictionary atom $\widehat{d}_{i,k}^{(t)}$ is a trivial task. But before using Theorem~\ref{thm:AllIteration}, we need to show that our assumption on $\varepsilon$ is indeed satisfied. In the following, we will prove that the assumption on $\varepsilon$ is satisfied if we perform $T_p$ power method and $T_c$ consensus iterations that are given according to the statement of Theorem~\ref{thm:main_result}.

\begin{IEEEproof}[Proof of Theorem~\ref{thm:main_result}]
After $T_d$ iterations of cloud $K$-SVD, error in any $k^{th}$ dictionary atom $\widehat{d}_{i,k}^{(T_d)}$ at site $i$ is a function of the error in $\widehat{E}_{i,k,R}^{(T_d)}$. Specifically, notice from~\eqref{eq:dkdkT} that we can write
\begin{align}\label{eqn:final_bound}
&\hspace{-2em}\|d_k^{(T_d)}d_k^{(T_d)^{\tT}}-\widehat{d}_{i,k}^{(T_d)}\widehat{d}_{i,k}^{(T_d)^{\tT}}\|_2 \nonumber\\
&\leq 8 N C_3\max_{i}{(\|B_{i,k,R}^{(T_d)}\|_2 C_4+ \|B_{i,k,R}^{(T_d)}\|_2^2)}+\varepsilon.
\end{align}
We can now upper bound $\|B_{i,k,R}^{(T_d)}\|_2$ in~\eqref{eqn:final_bound} using Theorem~\ref{thm:AllIteration}, but we first need to show that the statement of Theorem~\ref{thm:main_result} implies the assumption on $\varepsilon$ in Theorem~\ref{thm:AllIteration} is satisfied. That is, we need to show $\varepsilon \leq \frac{\delta_d}{8N\sqrt{n}C_3(1+\zeta)^{T_d-1}C_4(8C_3 NC_4^2+5)^{2(T_d K-2)}}$. Recall that by definition $\varepsilon = \mu {\nu}^{T_p}+4\epsilon^{3T_p}$. Substituting this, we must show that
{\small\begin{align*}
\label{eq:error1}
\mu {\nu}^{T_p}+4\epsilon^{3T_p}\leq \frac{\delta_d}{8N\sqrt{n}C_3(1+\zeta)^{T_d-1}C_4(8C_3 NC_4^2+5)^{2(T_d K-2)}}.
\end{align*}}
Since $\nu>0$ and $\epsilon>0$, therefore, $\mu {\nu}^{T_p}+4\epsilon^{3T_p}<
\mu (\nu + 4\epsilon^3)^{T_p}$. It is therefore sufficient to show that $\mu
({\nu} + 4\epsilon^3)^{T_p} \leq
\frac{\delta_d}{8N\sqrt{n}C_3(1+\zeta)^{T_d-1}C_4(8C_3 NC_4^2+5)^{2(T_d
K-2)}}$ for our selected values of $T_p$ and $T_c$. Showing that, however, is
a simple exercise and is left out for brevity. It therefore follows from
Theorem~\ref{thm:AllIteration} that
$\|d_k^{(T_d)}d_k^{(T_d)^{\tT}}-\widehat{d}_k^{(T_d)}\widehat{d}_k^{(T_d)^{\tT}}\|_2
\leq \varepsilon (1+\zeta)^{T_d-1} (8C_3 NC_4^2+5)^{(T_d-1)K+k-1}$.
{Substituting the upper bound on $\varepsilon$, we get
$\|d_k^{(T_d)}d_k^{(T_d)^{\tT}}-\widehat{d}_k^{(T_d)}\widehat{d}_k^{(T_d)^{\tT}}\|_2
\leq \delta_d$.}
\end{IEEEproof}

\section{Other results}\label{app:prior_results}
In this appendix, we collect some supporting results that are used in the proofs of our main results.

\begin{lemma}[Perturbation of singular values]\label{le:C_2_Assumption}
Let $D_2$ be a perturbed version of dictionary $D_1$ such that
$\|D_1-D_2\|_2\leq \epsilon_2$ and let $\Sigma_{T_0}$ be as defined in
Section~\ref{ssec:ksvd_prelim}. Then assuming
$\min_{\mathcal{I}\in\Sigma_{T_0}}\sigma_{T_0}\left(D_{1_{|\mathcal{I}}}\right)\geq
\sqrt{C_2^{\prime}} > \epsilon_2$, we have
 $\min_{\mathcal{I}\in\Sigma_{T_0}}\sigma_{T_0}\left(D_{2_{|\mathcal{I}}}\right)\geq \sqrt{C_2^{\prime}} - \epsilon_2$.
\end{lemma}
\begin{IEEEproof}
Using~\cite[Theorem~1]{stewart1998perturbation}, perturbation in $T_0^{th}$
singular value of $D_{1_{|\mathcal{I}}}$ can be bounded as
$|\sigma_{T_0}\left(D_{1_{|\mathcal{I}}}\right)-\sigma_{T_0}\left(D_{2_{|\mathcal{I}}}\right)|\leq
\|D_{1_{|\mathcal{I}}}-D_{2_{|\mathcal{I}}}\|_2\leq\|D_1-D_2\|_2\leq
\epsilon_2$. Using reverse triangular inequality, we therefor get $\forall
\mathcal{I}\in \Sigma_{T_0}, \epsilon_2 \geq
|\sigma_{T_0}\left(D_{1_{|\mathcal{I}}}\right)|-|\sigma_{T_0}\left(D_{2_{|\mathcal{I}}}\right)|\geq
\sqrt{C_2^{\prime}}-|\sigma_{T_0}\left(D_{2_{|\mathcal{I}}}\right)|$, which
leads to the claimed result.
\end{IEEEproof}

\begin{proposition}[Stability of sparse coding]\cite[Theorem~1]{Mehta.Gray.Conf2013}
\label{prop:SparseCodingBound}
Let $D_2$ be a perturbed version of dictionary $D_1$ such that $\|D_1-D_2\|_2\leq \epsilon_2$. Given any sample $y\in \mathbb{R}^{n}$, suppose sparse codes $x\in\mathbb{R}^{K}$ and $\widehat{x}\in \mathbb{R}^{K}$ are computed by solving the lasso problem~\eqref{eqn:lasso} using $D_1$ and $D_2$, respectively. Next, let $\min_{\substack{j\not\in \text{supp}(x)}}{\tau-|\langle d_{1,j}, y-D_1 x|>C_1}$, where $d_{1,j}$ denotes the $j^{th}$ atom of $D_1$, and suppose $D_1$ satisfies P2. Then, as long as $\epsilon_2\leq \frac{C_1^2 \tau}{44}$, we have that $\text{supp}(x)=\text{supp}(\widehat{x})$ and
{$\|x-\widehat{x}\|_2 \leq \frac{3\|D_1-D_2\|_2 \sqrt{T_0}}{\tau C_2},$}
where {$T_0=|\text{supp}(x)|$}.
\end{proposition}

Note that \cite[Theorem~1]{Mehta.Gray.Conf2013} also requires $D_2$ to
satisfy P2. Proposition~\ref{prop:SparseCodingBound} in its current form,
however, is a simple consequence of \cite[Theorem~1]{Mehta.Gray.Conf2013} and
Lemma~\ref{le:C_2_Assumption}.

\begin{proposition}[Perturbation of principal eigenvector]\cite[Chap. 8]{golub2012matrix}
\label{prop:PerturbedEigenvector} Let $A\in \mathbb{R}^{n\times n}$ be a
symmetric matrix and define $\widehat{A}=A+E$ to be a perturbed, but
symmetric version of $A$. Define $Q=\begin{bmatrix} q_1 & | & Q_2
\end{bmatrix}$ to be an orthogonal matrix comprising eigenvectors of $A$,
where $q_1$ denotes the principal eigenvector of $A$. Next, define $Q^{\tT}A
Q=\begin{bmatrix} \lambda & 0\\0 & \Lambda_2  \end{bmatrix}$ and
$Q^{\tT}EQ=\begin{bmatrix} \epsilon & e^{\tT}\\ e & E_{22} \end{bmatrix}$.
Then, using $\text{eig}(\Lambda_2)$ to denote the $(n-1)$ smallest
eigenvalues of $A$, it follows that if $g=\min_{\varrho\in
\text{eig}(\Lambda_2)}{|\lambda-\varrho|>0}$, and $\|E\|_F \leq \frac{g}{5}$
then there exists $p \in \mathbb{R}^{n-1}$ satisfying $\|p\|_2 \leq
\frac{4}{g}\|e\|_2$, such that $\widehat{q_1}=\frac{q_1+Q_2
p}{\sqrt{1+p^{\tT}p}}$ is a unit 2-norm principal eigenvector for
$\widehat{A}$. Moreover, $\|q_1 q_1^{\tT} - \widehat{q}_1
\widehat{q}_1^{\tT}\|_2 \leq \frac{4}{g}\|e\|_2.$
\end{proposition}

\begin{lemma}[Errors in vectors and their outerproducts]
\label{le:u_v}
For two unit $\ell_2$-norm vectors $u$ and $v$ if $\|uu^{\tT} - vv^{\tT}\|_2 \leq \epsilon$
and $u^{\tT}v \geq 0$ then $\|u-v\|_2\leq \sqrt{2}\epsilon.$
\end{lemma}
\begin{IEEEproof}
Let $\theta=\angle{(u,v)}$ and notice that $\|uu^{\tT}-vv^{\tT}\|_2 =
\sin\theta$. This implies $1-\cos^2{\theta}= \sin^2{\theta}
=\|uu^{\tT}-vv^{\tT}\|_2^2 \leq \epsilon^2$. Since $u$ and $v$ are unit norm
and $u^{\tT}v\geq 0$, we can write $\cos\theta=u^{\tT}v$. It then follows
that $ 1-u^{\tT}v \leq \frac{\epsilon^2}{1+u^{\tT}v} < \epsilon^2$. The claim
follows by noting that $\|u-v\|_2 = \sqrt{2(1-u^{\tT}v)}$.
\end{IEEEproof}
\end{appendices}